\documentclass[a4paper,11pt]{article}
\usepackage[top=4cm, bottom=4cm, left=3cm, right=3cm]{geometry}

\usepackage[utf8]{inputenc} 
\usepackage{hyperref}       
\usepackage{url}            
\usepackage{booktabs}       
\usepackage{amsfonts}       
\usepackage{nicefrac}       
\usepackage{microtype}      
\usepackage{amsmath,amsthm}
\usepackage{graphicx}

\usepackage{algorithm}

\newcommand{\Nystrom}[1]{{Nystr\"om}}

\providecommand{\scal}[2]{\left\langle{#1},{#2}\right\rangle}
\providecommand{\nor}[1]{\lVert{#1}\rVert}

\providecommand{\tr}{\operatorname{Tr}}

\newcommand{\R}{\mathbb R}

\newcommand{\N}{\mathbb N}

\newcommand{\hh}{\mathcal H}

\newcommand{\la}{\lambda}

\newcommand{\lspanc}[2]{\overline{\operatorname{span}\{#1~|~#2\}}}

\newcommand{\argmin}[1]{\mathop{\operatorname{argmin}}_{#1}}
\newcommand{\cond}[1]{{\operatorname{cond}{(#1)}}}

\newcommand{\X}{{X}}
\newcommand{\Y}{{\cal Y}}

\newcommand{\rhox}{{\rho_{\X}}}
\newcommand{\Ltwo}{{L^2(\X,\rhox)}}

\newcommand{\C}{{C}}

\newcommand{\K}{K_{nn}}
\newcommand{\Cn}{\widehat{\C}_n}
\newcommand{\Cl}{\C_\la}
\newcommand{\Cnl}{\widehat{\C}_{n\lambda}}
\newcommand{\Cm}{\widehat{\C}_M}
\newcommand{\Cml}{\widehat{\C}_{M\la}}
\newcommand{\Gm}{\widehat{G}_M}
\newcommand{\Gml}{\widehat{G}_{M\la}}
\newcommand{\Sm}{\widehat{S}_M}
\newcommand{\Sn}{\widehat{S}_n}

\newcommand{\Km}{K_{MM}}
\newcommand{\Knm}{K_{nM}}
\newcommand{\yn}{\widehat{y}}

\newcommand{\fh}{f_\hh}
\newcommand{\frho}{f_\rho}

\newcommand{\HS}{{\textrm{HS}}}

\newcommand{\iter}{t}

\newcommand{\eqals}[1]{\begin{align*}#1\end{align*}}
\newcommand{\eqal}[1]{\begin{align}#1\end{align}}
\newcommand{\bpr}{\begin{proof}}
\newcommand{\epr}{\end{proof}}
\newcommand{\be}{\begin{equation}}
\newcommand{\ee}{\end{equation}}

\newtheorem{definition}{Definition}
\newcommand{\bd}{\begin{definition}}
\newcommand{\ed}{\end{definition}}

\newcommand{\bi}{\begin{itemize}}
\newcommand{\ei}{\end{itemize}}

\newtheorem{ass}{Assumption}
\newcommand{\ba}{\begin{ass}}
\newcommand{\ea}{\end{ass}}

\newtheorem{remark}{Remark}
\newcommand{\br}{\begin{remark}}
\newcommand{\er}{\end{remark}}

\newtheorem{example}{Example}
\newcommand{\bex}{\begin{example}}
\newcommand{\eex}{\end{example}}

\newtheorem{proposition}{Proposition}
\newcommand{\bp}{\begin{proposition}}
\newcommand{\ep}{\end{proposition}}

\newtheorem{lemma}{Lemma}
\newcommand{\blm}{\begin{lemma}}
\newcommand{\elm}{\end{lemma}}

\newtheorem{theorem}{Theorem}
\newcommand{\bt}{\begin{theorem}}
\newcommand{\et}{\end{theorem}}

\newtheorem{corollary}{Corollary}
\newcommand{\bcor}{\begin{corollary}}
\newcommand{\ecor}{\end{corollary}}

\newcommand{\loz}[1]{{#1}}

\title{FALKON:  An Optimal Large Scale Kernel Method}

\author{
	Alessandro Rudi \thanks{E-mail: \texttt{alessandro.rudi@inria.fr}. This work was done when A.R. was working at Laboratory of Computational and Statistical Learning (Istituto Italiano di Tecnologia).}  \\
	{\small INRIA -- Sierra Project-team,} \\
	{\small École Normale Supérieure, Paris} \\
	\and
	Luigi Carratino\\
	{\small University of Genoa} \\
	{\small Genova, Italy}\\
	\and
	Lorenzo Rosasco \\
	{\small University of Genoa,} \\
	{\small LCSL, IIT \& MIT} \\
}

\begin{document}
	
	\maketitle

	\begin{abstract}
		Kernel methods provide a principled way to perform non linear, nonparametric learning.
		They rely on solid functional analytic foundations and enjoy optimal statistical properties.
		However, at least in their basic form,  they have limited  applicability in large scale scenarios because of stringent computational requirements  in terms of time and especially memory. In this paper, we take a substantial step in scaling up kernel methods, proposing FALKON, a novel algorithm that allows to efficiently process millions of points. FALKON is derived combining several algorithmic principles, namely stochastic subsampling, iterative solvers and preconditioning. Our theoretical analysis shows that  optimal statistical accuracy  is achieved  requiring essentially $O(n)$ memory and $O(n\sqrt{n})$  time. \loz{An extensive experimental analysis on large scale datasets shows that, even with a single machine,  FALKON   outperforms  previous state of the art solutions, which exploit parallel/distributed architectures.}
	\end{abstract}
	
	\section{Introduction}

	The goal in supervised learning  is to learn  from examples a function that predicts well new data. Nonparametric methods are often crucial since the functions to be learned can be  non-linear and complex Kernel methods are probably the most popular among nonparametric learning methods, but despite excellent theoretical properties, they have limited applications in large scale learning because of time and memory requirements,  typically at least quadratic in the number of data points. \\
	Overcoming these limitations  has motivated a variety of practical approaches including gradient methods,  as well  accelerated, stochastic and preconditioned extensions, to improve time complexity \cite{CapYao06,journals/neco/GerfoROVV08,fasshauer2012stable,avron2016faster, gonen2016solving, ma2017diving}. Random projections provide  an approach to reduce memory requirements,
	popular methods including \Nystrom{} \cite{conf/nips/WilliamsS00,conf/icml/SmolaS00}, random features \cite{conf/nips/RahimiR07},  and  their numerous extensions.  From a theoretical perspective a key question
	has become to characterize statistical and computational trade-offs, that is if,  or under which conditions, computational gains come at the expense of statistical accuracy. In particular, recent results considering least squares,  show that there are large class of problems for which, by combining \Nystrom{} or random features approaches \cite{rahimi2009weighted, conf/colt/Bach13,alaoui2014fast,rudi2015less,rudi2016rf,bach17quadrature} with ridge regression, it is possible to substantially reduce computations, while preserving the same optimal statistical accuracy of  exact kernel ridge regression (KRR). While  statistical lower bounds exist for this setting, there are no corresponding computational lower bounds.  The state of the art approximation of KRR, for which optimal statistical bounds are known, typically requires complexities that are  roughly $O(n^2)$ in time  \loz{and  memory (or possibly $O(n)$ in memory,  if kernel computations are made on the fly). }
	
	\loz{In this paper, we propose and study FALKON,   a new algorithm that, to the best of our knowledge,  has the best known theoretical guarantees.  At the same time FALKON provides an efficient approach to apply kernel methods on millions of points, and  tested on a variety of large scale problems outperform previously proposed methods while utilizing only a fraction of computational resources.}
	More precisely,  we take a substantial step in provably reducing the computational requirements,  showing that,  up to logarithmic factors, a time/memory complexity of $O(n\sqrt{n})$ and $O(n)$ is sufficient for optimal statistical accuracy. Our new algorithm, exploits the idea of using \Nystrom{} methods to approximate the KRR problem, but also to efficiently compute a preconditioning to be used in conjugate gradient. To the best of our knowledge this is the first time all these ideas are combined and put to fruition. Our theoretical analysis derives optimal statistical rates both in a basic setting and under benign conditions for which fast rates are possible. The potential benefits of different sampling strategies are also analyzed. Most importantly,  the empirical performances  are thoroughly tested on available large  scale data-sets. Our results show that, even on a single machine, FALKON can outperforms  state of the art methods  on most problems  both in terms of time efficiency and prediction accuracy.
	\loz{In particular, our results suggest that FALKON could  be a viable kernel alternative to deep fully connected neural networks  for large scale problems.
		
		The rest of the paper is organized as follows. In Sect.~\ref{sect:background} we give some background on kernel methods.  In Sect.~\ref{sect:FALKON} we introduce FALKON,  while in  Sect.~\ref{sect:teo} we present and discuss the main technical results. Finally in Sect.~\ref{sect:experiments} we present  experimental results. 
	}
	\section{Statistical and Computational Trade-offs \\ in Kernel Methods}\label{sect:background}
	We consider the supervised learning problem of estimating a function from random noisy samples. In statistical learning theory, this can be  formalized as the problem of solving
	\be\label{eq:exrm} \inf_{f \in \hh} {\cal E}(f), \quad\quad {\cal E}(f) = \int (f(x)-y)^2 d\rho(x,y),\ee
	given samples $(x_i, y_i)_{i=1}^n$ from $\rho$, which is fixed but unknown and where, $\hh$ is a space of candidate solutions.  Ideally, a good empirical solution  $\widehat{f}$ should have small {\em excess risk}
	\eqal{\label{eq:excess-risk}
		{\cal R}(\widehat{f}~) ~~ = ~~ {\cal E}(\widehat{f}~) ~ - \inf_{f \in \hh} {\cal E}(f),
	}
	since this implies it will generalize/predict well new data. In this paper, we are interested in both computational and statistical aspects of the above problem. In particular, we investigate the computational resources needed to achieve optimal statistical accuracy, i.e. minimal excess risk.
	Our focus is on the most popular class of nonparametric methods, namely kernel methods.
	
	\paragraph{Kernel methods and ridge regression.} Kernel methods consider  a space  $\hh$ of  functions
	\eqal{\label{eq:basic-model}
		f(x) = \sum_{i=1}^n \alpha_j K(x, x_i),
	}
	where $K$ is a positive definite kernel \footnote{$K$ is positive definite, if  the matrix  with entries  $K(x_i, x_j)$ is positive semidefinite  $\forall x_1, \dots, x_N, N \in \N$~ \cite{schlkopf2002learning}}. The coefficients $\alpha_1, \dots,\alpha_n$ are typically derived from a convex optimization problem, that  for the square loss  is
	\eqal{\label{eq:reg-emp-risk}
		\widehat{f}_{n,\la} = \argmin{f \in \hh} \frac{1}{n} \sum_{i=1}^n (f(x_i)-y_i)^2 + \la \nor{f}^2_\hh,
	}
	and defines the so called kernel ridge regression (KRR) estimator \cite{schlkopf2002learning}. An advantage of least squares approaches is that they reduce computations to  a linear system
	\eqal{\label{eq:krls}
		(\K + \la n I)~\alpha = \yn,
	}
	where $\K$ is an $n\times n$ matrix defined by $(\K)_{ij} = K(x_i, x_j)$ and $\yn = (y_1, \dots y_n)$.
	We next comment on  computational and statistical  properties of KRR.\\
	
	\noindent{\em Computations}.~ 
	Solving Eq.~\eqref{eq:krls} for  large datasets is  challenging. A  direct approach  requires $O(n^2)$ in space, to allocate  $\K$, $O(n^2)$ kernel evaluations, and $O(n^2 c_K + n^3)$ in time, to compute  and invert $\K$ ($c_K$ is the kernel evaluation cost assumed constant and omitted throughout).\\
	
	\noindent{\em Statistics}.~  Under basic assumptions, KRR achieves an error
	$ {\cal R}(\widehat{f}_{\la_n}) = O(n^{-1/2}),$
	for $\la_n = n^{-1/2}$, which is optimal in a minimax sense and can be  improved {\em only} under more stringent assumptions  \cite{caponnetto2007optimal,steinwart2009optimal}.
	
	The  question is then if it is  possible to achieve the statistical properties of KRR, with less computations.
	
	\paragraph{Gradient methods and early stopping.} A natural idea is to consider iterative solvers and in particular gradient methods, because of their simplicity and low iteration cost. A basic example is computing the coefficients in~\eqref{eq:basic-model} by
	\be\label{eq:it}
	\alpha_t = \alpha_{t-1} + \tau \left[(\K \alpha_{t-1} - \yn) + \la n \alpha_{t-1}\right],
	\ee
	for a suitable step-size choice  $\tau$. \\
	
	\noindent{\em Computations}.~
	In this case, if $t$ is  the number of iterations, gradient methods require $O(n^2 t)$ in time, $O(n^2)$ in memory and $O(n^2)$ in kernel evaluations, if the kernel matrix is stored.  Note that,  the kernel matrix can  also be computed on the fly with  only $O(n)$ memory,  but $O(n^2 t)$ kernel evaluations are   required.
	We note that, beyond the above simple iteration, several variants have been considered including accelerated \cite{CapYao06,bauer} and stochastic extensions \cite{dieuleveut2014non}.\\
	
	\noindent{\em Statistics}.~
	The statistical properties of iterative approaches are  well studied and also in the case where $\la$ is set to zero, and regularization is performed by choosing a suitable stopping time \cite{yao}.
	In this latter case, the number of iterations can roughly be thought of $1/\la$ and $O(\sqrt{n})$ iterations are needed for basic gradient descent,  $O(n^{1/4})$ for accelerated methods and possible $O(1)$ iterations/epochs for stochastic methods.
	Importantly,  we note that unlike most optimization studies, here we are considering the number of iterations needed to  solve~\eqref{eq:exrm},  rather than~\eqref{eq:reg-emp-risk}. 
	
	While the time complexity of these methods  dramatically improves over KRR, and computations can be done in blocks, memory requirements (or number of kernel evaluations) still makes the application to large scale setting cumbersome. Randomization provides an approach to tackle this challenge.
	%
	%
	%
	%
	%
	%
	
	\paragraph{Random projections.} The rough idea is to use random projections  to compute  $\K$ only approximately. The most popular examples in this class of approaches are  { \Nystrom{}} \cite{conf/nips/WilliamsS00,conf/icml/SmolaS00} and { random features} \cite{conf/nips/RahimiR07} methods. In the following we focus in particular on a basic \Nystrom{} approach based on considering functions of the form
	\eqal{\label{eq:form-subsampled}
		\widetilde{f}_{\la, M}(x) = \sum_{i=1}^M \widetilde{\alpha}_i K(x,\widetilde{x}_i), \quad \textrm{with} \quad \{\widetilde{x}_1,\dots, \widetilde{x}_M\} \subseteq \{x_1,\dots,x_n\},
	}
	defined considering only a subset of $M$ training points sampled uniformly. In this case, there are only $M$ coefficients that, following the approach in~\eqref{eq:reg-emp-risk}, can be derived considering  the linear system
	\eqal{\label{eq:base-nystrom}
		H \widetilde{\alpha} ~=~ z,\quad\quad \textrm{where}  \quad H = K_{nM}^\top K_{nM} ~+~ \lambda n K_{MM}, \quad z = K_{nM}^\top \hat{y}.
	}
	Here $\Knm$ is the  $n \times M$ matrix with  $(\Knm)_{ij} = K(x_i,\widetilde{x}_j)$ and $\Km$ is the $M\times M$ matrix with $(\Km)_{ij} = K(\widetilde{x}_i,\widetilde{x}_j)$. This method consists in subsampling the columns of $\K$ and can be seen as a particular form of random projections.\\ 
	
	\noindent{\em Computations}.~ Direct methods for solving~\eqref{eq:base-nystrom} require $O(n M^2)$ in time to form $\Knm^\top \Knm$ and $O(M^3)$ for solving the linear system, and  only $O(nM)$ kernel evaluations.  The naive memory requirement is $O(nM)$ to store $\Knm$, however if $\Knm^\top \Knm$ is computed in blocks  of dimension at most $M \times M$ only $O(M^2)$ memory is needed.
	Iterative approaches  as in~\eqref{eq:it} can also be  combined with  random projections \cite{dai2014scalable,camoriano2016nytro,tu2016large} to slightly reduce time requirements (see Table.~\ref{tab:complexities}, or Sect.~\ref{sect:long-comparison} in the appendix, for more details).\\
	
	\noindent{\em Statistics}.~ The key point though, is that random projections allow to dramatically reduce memory requirements as soon as $M\ll n$ and the question arises of whether  this comes at expenses of statistical accuracy. Interestingly, recent results considering this question show that there  are large classes of problems for which
	$M=\tilde O(\sqrt{n})$ suffices for the same optimal statistical accuracy of the exact KRR \cite{conf/colt/Bach13,alaoui2014fast,rudi2015less}. 
	
	In summary,  in this case the computations needed for optimal statistical accuracy are   reduced from $O(n^2)$ to $O(n\sqrt{n})$ kernel evaluations, but the best time complexity is basically  $O(n^2)$. In the rest of the paper we discuss how this requirement can indeed be dramatically reduced.

	\section{FALKON}\label{sect:FALKON}
	Our approach is based on a novel combination of  randomized projections with iterative solvers plus preconditioning.  The main novelty is that we use random projections to approximate both  the problem {\em and}  the preconditioning.
	
	\paragraph{\loz{Preliminaries: preconditioning and KRR.}} We begin recalling the basic idea behind preconditioning. The key quantity is the condition number,  that for a linear system is the ratio between the largest and smallest singular values of the matrix defining the problem  \cite{saad2003iterative}. For example, for problem~\eqref{eq:krls} the condition number is given by
	$$\text{cond}(\K+\la n I )= (\sigma_{\text{max}}+\la n)/(\sigma_{\text{min}}+\la n),$$
	with $\sigma_{\text{max}},\sigma_{\text{min}}$ largest and smallest  eigenvalues  of $\K$, respectively. The importance of the condition number is that it captures the time complexity of iteratively solving the corresponding linear system. For example, if a simple gradient descent~\eqref{eq:it} is used,  the number of iterations needed for an $\epsilon$ accurate solution of problem~\eqref{eq:krls}  is   
	$$t = O(\text{cond}(\K+\la n I) \log(1/\epsilon)).$$
	It is shown in \cite{camoriano2016nytro} that in this case $t=\sqrt{n}\log n$ are needed   to achieve a solution with good statistical properties. Indeed, it can be shown that
	roughly $t \approx 1/\la \log(\frac{1}{\epsilon})$ are needed where  $\la=1/\sqrt{n}$ and $\epsilon=1/n$.
	The idea behind preconditioning is  to use a suitable  matrix $B$ to define an equivalent  linear system with better condition number.
	For~\eqref{eq:krls}, an ideal choice is  $B$ such that
	\be\label{eq:precK}
	B B^\top = (\K+\la n I)^{-1}
	\ee
	and  $B^\top (\K+\la n I) B ~ \beta ~=~ B^\top \hat y.$ Clearly, if  $\beta_*$ solves  the latter problem,  $\alpha_*=B \beta_*$ is a solution of problem~\eqref{eq:krls}. Using a   preconditioner $B$  as in~\eqref{eq:precK} one iteration is sufficient, but  computing the $B$ is typically as hard as the original problem.
	The problem is  to derive preconditioning such that~\eqref{eq:precK} might hold only approximately, but that  can be computed efficiently.
	Derivation of efficient preconditioners for the exact KRR problem~\eqref{eq:krls} has been the subject of recent studies, \cite{fasshauer2012stable,avron2016faster, cutajar2016preconditioning, gonen2016solving, ma2017diving}. In particular, \cite{avron2016faster, cutajar2016preconditioning, gonen2016solving, ma2017diving} consider  random projections to approximately compute a preconditioner. Clearly, while preconditioning~\eqref{eq:krls} leads to computational speed ups in terms of the number of iterations,  requirements in terms of memory/kernel evaluation are  the same as standard kernel ridge regression. 
	
	\loz{The key idea  to tackle this problem is to consider an efficient  preconditioning approach for problem~\eqref{eq:base-nystrom} rather than~\eqref{eq:krls}.}
	%
	
	\paragraph{Basic FALKON algorithm.} We begin illustrating a basic version of our approach. The key ingredient is the
	following preconditioner for  Eq.~\eqref{eq:base-nystrom},
	\eqal{\label{eq:B-base}
		B B^\top = \left(\frac{n}{M}\Km^2 + \la n \Km\right)^{-1},
	}
	which is itself based on a \Nystrom{} approximation\footnote{
		For the sake of simplicity, here we assume $\Km$ to be invertible and the \Nystrom{} centers selected with uniform sampling from the training set, see Sect.~\ref{sect:gen-algo} and Alg.~\ref{algo:main-MATLAB} in the appendix for the general algorithm.}.
	The above preconditioning  is a natural approximation of the ideal preconditioning of problem~\eqref{eq:base-nystrom} that corresponds to $B B^\top = (K_{nM}^\top K_{nM} ~+~ \lambda n K_{MM})^{-1}$ and reduces  to it if $M=n$.
	Our  theoretical analysis, shows that $M\ll n$ suffices for deriving optimal statistical rates.  In its basic form  FALKON is derived
	combining  the above preconditioning and   gradient descent,
	\eqal{\label{eq:simple-FALKON}
		\widehat{f}_{\la, M, \iter}(x) &= \sum_{i=1}^M \alpha_{\iter,i}K(x,\widetilde{x}_i), \quad \textrm{with} \quad \alpha_{\iter} = B \beta_{\iter} \quad \textrm{and} \\
		\beta_k &= \beta_{k-1} - \frac{\tau}{n} B^\top \left[\Knm^\top(\Knm (B \beta_{k-1})  -  \yn) ~+~  \lambda n \Km (B \beta_{k-1})\right],
	}
	for $\iter \in \N$, $\beta_0 = 0$ and $1 \leq k \leq \iter$ and a suitable chosen $\tau$. In practice, a refined version of FALKON is preferable where a faster gradient iteration  is used and  additional care is taken in organizing computations.
	\begin{algorithm}[t]
		\caption{{\tt MATLAB} code for FALKON. It requires $O(nMt + M^3)$ in time and $O(M^2)$ in  memory. See Sect.~\ref{sect:gen-algo} and Alg.~\ref{algo:main-MATLAB} in the appendixes for the complete algorithm.
			\label{algo:main}}
		\vspace{0.1cm}
		{\small
			\begin{flushleft}
				{\bf Input:} Dataset $X = (x_i)_{i=1}^n \in \R^{n\times D}, \hat y = (y_i)_{i=1}^n \in \R^{n}$, centers $C = (\tilde{x}_j)_{j=1}^M \in \R^{M \times D}$, $\text{KernelMatrix}$ computing the kernel matrix given two sets of points, regularization parameter $\la$, number of iterations $\iter$.\\
				{\bf Output:} \Nystrom{} coefficients $\alpha$.
			\end{flushleft}
			\begin{center}
				\begin{verbatim}
				function alpha = FALKON(X, C, Y, KernelMatrix, lambda, t)
				n = size(X,1); M = size(C,1); KMM = KernelMatrix(C,C);
				T = chol(KMM + eps*M*eye(M));
				A = chol(T*T'/M + lambda*eye(M));
				
				function w = KnM_times_vector(u, v)
				w = zeros(M,1); ms = ceil(linspace(0, n, ceil(n/M)+1));
				for i=1:ceil(n/M)
				Kr = KernelMatrix( X(ms(i)+1:ms(i+1),:), C );
				w = w + Kr'*(Kr*u + v(ms(i)+1:ms(i+1),:));
				end
				end
				
				BHB = @(u) A'\(T'\(KnM_times_vector(T\(A\u), zeros(n,1))/n) + lambda*(A\u));
				r = A'\(T'\KnM_times_vector(zeros(M,1), Y/n));
				alpha = T\(A\conjgrad(BHB, r, t));
				end
				\end{verbatim}
			\end{center}
		}
		\vspace{-0.35cm}
	\end{algorithm}
	
	\paragraph{FALKON.} The actual version of FALKON we propose is Alg.~\ref{algo:main} (see Sect.~\ref{sect:gen-algo}, Alg.~\ref{algo:main-MATLAB} for the complete algorithm). It  consists in solving the system $B^\top H B \beta = B^\top z$ via conjugate gradient \cite{saad2003iterative}, since it is a fast gradient method and does not require to specify the  step-size. Moreover, to compute $B$ quickly, with reduced numerical errors, we consider the following strategy
	\eqal{\label{eq:def-P}
		B = \frac{1}{\sqrt{n}}T^{-1}A^{-1}, \qquad T = \textrm{chol}(K_{MM}),\quad A = \textrm{chol}\left( \frac{1}{M}~T\,T^\top + \lambda I \right),
	}
	where $\textrm{chol}()$ is the Cholesky decomposition (in Sect.~\ref{sect:gen-algo} the strategy for non invertible $K_{MM}$).\\
	
	\noindent{\em Computations}.~ in Alg.~\ref{algo:main}, $B$ is never built explicitly and $A, T$ are two upper-triangular matrices, so $A^{-\top}u, A^{-1}u$ for a vector $u$ costs $M^2$, and the same for $T$. The cost of computing the preconditioner is only $\frac{4}{3} M^3$ floating point operations (consisting in two Cholesky decompositions and one product of two triangular matrices). Then FALKON requires $O(nMt + M^3)$ in time and the same $O(M^2)$ memory requirement of the basic \Nystrom{} method,  if  matrix/vector multiplications at each iteration are performed in  blocks. This implies $O(nMt)$  kernel evaluations are needed. 
	%
	
	The question  remains to characterize $M$ and the number of iterations needed for good statistical accuracy. 
	\loz{Indeed, in the next section we show that roughly $O(n\sqrt{n})$ computations and $O(n)$ memory are sufficient for optimal accuracy. This implies that FALKON  is currently  the most efficient kernel method with the  same optimal statistical accuracy of KRR, see Table~\ref{tab:complexities}.}


	%
	%
	%
	%
	%
	
	%
	%
	%
	%

	%


	%
	\section{Theoretical Analysis}\label{sect:teo}
	
	In this section,  we characterize the generalization properties of FALKON showing it  achieves the  optimal generalization error of KRR, with dramatically reduced computations. This result is given in Thm.~\ref{thm:simple-rates} and derived in two steps. First, we study the difference between the excess risk of FALKON and that of the basic \Nystrom{}~\eqref{eq:base-nystrom},  showing it depends on the condition number induced by the preconditioning,  hence on $M$ (see Thm.\ref{thm:oracle-inequality-base}).  \loz{Deriving these results requires some care, since differently to standard optimization results, our goal is to solve~\eqref{eq:exrm} i.e.  achieve small excess risk, not to minimize the empirical error.} 
	Second, we show that choosing $M = \widetilde{O}(1/\la)$ allows to make this difference as  small as $e^{-\iter/2}$  (see Thm.\ref{thm:M-simple-rates}).  Finally, recalling that the basic \Nystrom{} for $\la=1/\sqrt{n}$ has essentially the same statistical properties of KRR \cite{rudi2015less}, we  answer the question posed at the end of the last section  and show that roughly
	$\log n$ iterations are sufficient for optimal statistical accuracy. Following the discussion in the previous section this means that the
	computational requirements for optimal accuracy are $\widetilde{O}(n \sqrt{n})$ in time/kernel evaluations and $\widetilde{O}(n)$ in space. Later in this section faster rates under further regularity assumptions are also derived and the effect of different selection methods for the \Nystrom{} centers  considered. The proofs for this section are provided in Sect.~\ref{sect:proof-main-res} of the appendixes.
	
	
	%
	
	\subsection{Main Result}
	\loz{The first result  is interesting in its own right since  it corresponds to translating optimization guarantees into statistical results.} In particular, we  derive a  relation the excess risk  of the FALKON algorithm {\small $\widehat{f}_{\la,M, t}$} from Alg.~\ref{algo:main}
	and the  \Nystrom{} estimator {\small $\widetilde{f}_{\la, M}$} from Eq.~\eqref{eq:base-nystrom} with uniform sampling.
	
	\bt\label{thm:oracle-inequality-base}
	Let $n, M \geq 3$, $\iter \in \N$, $0 < \la \leq \la_1$ and $\delta \in (0,1]$. Assume  there exists $\kappa \geq 1$ such that $K(x,x) \leq \kappa^2$ for any $x \in \X$. Then,  the following inequality holds with probability $1 - \delta$
	$$
	{\cal R}(\widehat{f}_{\la,M,t})^{1/2} ~~\leq~~ {\cal R}(\widetilde{f}_{\la, M})^{1/2} ~~ + ~~ 4\widehat{v} ~ e^{- \nu \iter} ~ \sqrt{1 + \frac{9\kappa^2}{\la n} \log \frac{n}{\delta}},
	$$
	
	where $\widehat{v}^2 = \frac{1}{n} \sum_{i=1}^n y_i^2$ and $\nu = \log(1 + 2/(\cond{B^\top H B}^{1/2}- 1))$, with $\cond{B^\top H B}$ the condition number of $B^\top H B$. Note that $\la_1 > 0$ is a constant not depending on $\la, n, M, \delta, \iter$.
	\et
	The additive term in the bound above decreases exponentially in the number of iterations. If the condition number of $B^\top H B$ is smaller than  a small universal constant (e.g. $17$), then $\nu > 1/2$ and  the additive term decreases as $e^{-\frac{\iter}{2}}$.  
	Next,  theorems derive a condition on $M$ that allows to control $\cond{B^\top H B}$, and  derive such an  exponential decay.
	\bt\label{thm:M-simple-rates}
	Under the same conditions of Thm.~\ref{thm:oracle-inequality-base}, if
	$$
	M \geq 5\left[1 + \frac{14\kappa^2}{\la}\right]\log \frac{8\kappa^2}{\la \delta}.
	$$
	then the exponent $\nu$ in Thm.~\ref{thm:oracle-inequality-base} satisfies $\nu \geq 1/2$.
	\et
	The above result gives  the desired exponential bound showing that after $\log n$ iterations
	the excess risk of  FALKON is controlled by that of the basic \Nystrom{}, more precisely
	$${\cal R}(\widehat{f}_{\la,M,\iter}) \leq 2 {\cal R}(\widetilde{f}_{\la, M}) \quad \textrm{when} \quad \iter \geq \log {\cal R}(\widetilde{f}_{\la, M}) ~+~ \log \left(1 + \frac{9\kappa^2}{\la n} \log \frac{n}{\delta}\right) ~+~ \log \left(16\widehat{v}^2\right).$$
	\begin{table}
		\centering
		\resizebox{\textwidth}{!}{%
\begin{tabular}{l  c c c c}
\toprule 
Algorithm & train time & kernel evaluations & memory  & test time \\
\midrule
SVM / KRR + direct method & $n^3$ & $n^2$ & $n^2$ & $n$\\
KRR + iterative \cite{CapYao06,journals/neco/GerfoROVV08} & $n^2\sqrt[4]{n}$ & $n^2$ & $n^2$ & $n$ \\
Doubly stochastic \cite{dai2014scalable} & $n^{2}\sqrt{n}$ & $n^{2}\sqrt{n}$ & $n$ & $n$\\ 
Pegasos / KRR + sgd \cite{shalev2011pegasos} & $n^{2}$ & $n^{2}$ & $n$ & $n$ \\
KRR + iter + precond \cite{fasshauer2012stable,yang2015randomized,avron2016faster, gonen2016solving, ma2017diving} & $n^2$ & $n^2$ & $n$  & $n$ \\ 
Divide \& Conquer \cite{conf/colt/ZhangDW13} & $n^2$ & $n\sqrt{n}$ & $n$ & $n$\\
\Nystrom{}, random features \cite{conf/nips/WilliamsS00,conf/icml/SmolaS00,conf/nips/RahimiR07}  & $n^2$ & $n\sqrt{n}$ & $n$ & $\sqrt{n}$\\
\Nystrom{} + iterative \cite{camoriano2016nytro,tu2016large} & $n^2$ & $n\sqrt{n}$ & $n$ & $\sqrt{n}$\\
\Nystrom{} + sgd \cite{dieuleveut2014non} & $n^2$ & $n\sqrt{n}$ & $n$ & $\sqrt{n}$\\
{\bf FALKON} (see Thm.~\ref{thm:simple-rates}) & $\boldsymbol{n\sqrt{n}}$ & $\boldsymbol{n\sqrt{n}}$ & $\boldsymbol{n}$ & $\boldsymbol{\sqrt{n}}$\\
\bottomrule
\end{tabular}
}
		\caption{Computational complexity required by different algorithms, for optimal generalization. Logarithmic terms are not showed. \label{tab:complexities}}
	\end{table}
	
	Finally, we derive an excess risk bound for FALKON. By the no-free-lunch theorem, this requires some conditions on the learning problem.
	We first consider a standard basic setting where we only assume it exists $\fh \in \hh$ such that ${\cal E}(\fh) = \inf_{f\in \hh} {\cal E}(f)$.
	\bt\label{thm:simple-rates}
	Let $\delta \in (0,1]$. \loz{Assume  there exists $\kappa \geq 1$ such that $K(x,x) \leq \kappa^2$ for any $x \in \X$, and} $y \in [-\frac{a}{2}, \frac{a}{2}]$, almost surely,  $a > 0$. There exist  $n_0 \in \N$ such that for any $n \geq n_0$,  if
	$$\la = \frac{1}{\sqrt{n}}, \qquad M \geq 75 ~ \sqrt{n} ~ \log\frac{48\kappa^2 n}{ \delta}, \qquad \iter ~\geq~ \frac{1}{2}\log(n) ~+~ 5 + 2\log (a + 3\kappa),$$
	then with probability $1-\delta$,
	$$ {\cal R}(\widehat{f}_{\la, M, t}~) \leq \frac{c_0 \log^2 \frac{24}{\delta}}{\sqrt{n}}.$$
	In particular $n_0, c_0$ do not depend on $\la, M, n, t$ and $c_0$ do not depend on $\delta$.
	\et
	The above result provides the desired bound, and all the constants  are given in the appendix. The obtained learning rate is the same as the full KRR estimator  and is known to be optimal in a minmax sense \cite{caponnetto2007optimal}, hence not improvable. As mentioned before,  the same bound is also achieved by the basic \Nystrom{} method but with much worse time complexity. Indeed, as discussed before,  using a simple iterative solver  typically requires $O(\sqrt{n}\log n)$ iterations, \loz{while we need only $O(\log n)$. Considering the choice for $M$ this leads to a computational time of $O(nMt) = O(n\sqrt{n})$ for optimal generalization (omitting logarithmic terms)}. To the best of our knowledge FALKON currently provides the best time/space complexity to achieve the statistical accuracy of KRR.
	Beyond the basic setting considered above, in the next section we show that FALKON can achieve much faster rates under refined regularity assumptions and also consider the potential benefits of leverage score sampling.
	\subsection{Fast learning rates and \Nystrom{} with approximate leverage scores}\label{sect:fast-rates}
	Considering fast rates and  \Nystrom{} with more general sampling is considerably more technical and a heavier notation is needed.
	Our analysis  apply to any approximation scheme (e.g. \cite{journals/jmlr/DrineasMMW12,alaoui2014fast,conf/innovations/CohenLMMPS15}) satisfying the definition of $q$-approximate leverage scores \cite{rudi2015less}, satisfying
	$q^{-1} l_i(\la) \leq \widehat{l}_i(\la) \leq ql_i(\la), \quad \forall ~i \in \{1,\dots, n\}.$
	Here $\la > 0$, $l_i(\la) = (\K(\K + \la n I)^{-1})_{ii}$ are the leverage scores and $q \geq 1$ controls the quality of the approximation. In particular, given $\la$, the \Nystrom{} points are sampled independently from the dataset with probability $p_i \propto \widehat{l}_i(\la)$.
	We need a few more definitions.  Let  $K_x=K(x, \cdot)$ for any $x \in \X$ and  $\hh$ the reproducing kernel Hilbert space \cite{steinwart2008support} of functions with inner product defined by $\hh = \lspanc{K_x}{x \in \X}$ and closed with respect to the inner product $\scal{\cdot}{\cdot}_\hh$ defined by $\scal{K_x}{K_{x'}}_\hh = K(x,x')$, for all $x,x' \in \X$. Define $C: \hh \to \hh$ to be the linear operator
	$\scal{f}{C g}_\hh = \int_{\X} f(x) g(x) d\rhox(x)$, for all $f,g \in \hh$.
	Finally define the following quantities,
	$$ {\cal N}_{\infty}(\la) = \sup_{x \in \X} \nor{(C + \la I)^{-1/2} K_x}_{\hh}, \quad {\cal N}(\la) = \tr(C(C+\la I)^{-1}).$$
	The latter quantity is known as {\em degrees of freedom} or {\em effective dimension}, can be seen as a measure of the {\em size}  of $\hh$.
	The quantity ${\cal N}_\infty(\la)$ can be seen to provide a uniform bound on the leverage scores. In particular note that ${\cal N}(\la) \leq {\cal N}_{\infty}(\la) \leq \frac{\kappa^2}{\la}$ \cite{rudi2015less}.
	We can now  provide a refined version of Thm.~\ref{thm:M-simple-rates}.
	%
	\bt\label{thm:M-fast-rates}
	Under the same conditions of Thm.~\ref{thm:oracle-inequality-base}, the exponent $\nu$ in Thm.~\ref{thm:oracle-inequality-base} satisfies $\nu \geq 1/2$, when \\
	${}\quad 1.$ either \Nystrom{} uniform sampling is used with $M \geq 70\left[1 + {\cal N}_\infty(\la)\right]\log \frac{8\kappa^2}{\la \delta}.$ \\
	${}\quad 2.$ or \Nystrom{} $q$-approx. lev. scores \cite{rudi2015less} is used, with $\la \geq \frac{19\kappa^2}{n} \log \frac{n}{2\delta}$, $n \geq 405 \kappa^2 \log \frac{12\kappa^2}{\delta},$
	$$
	M \geq 215\left[2 + q^2{\cal N}(\la)\right]\log \frac{8\kappa^2}{\la \delta}.
	$$
	\et
	\begin{figure}[t]
		\begin{center}
			\includegraphics[width=0.7\linewidth]{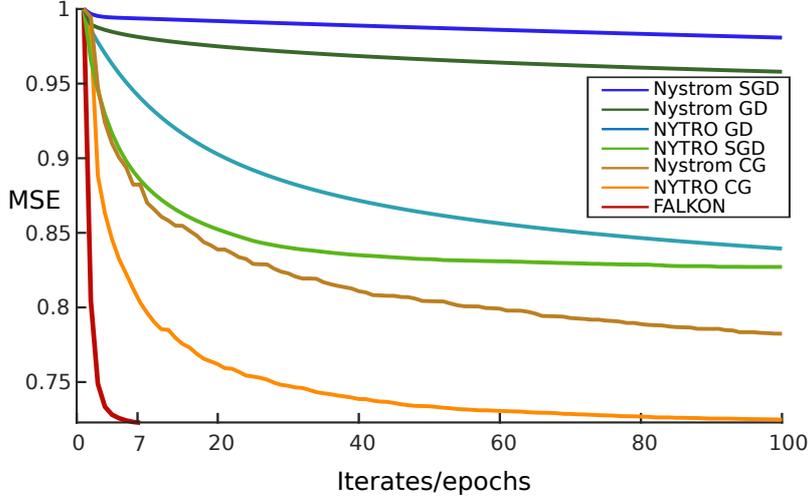}
		\end{center}
		\caption{Falkon is compared to stochastic gradient, gradient descent and conjugate gradient applied to Problem~\eqref{eq:base-nystrom}, while NYTRO refer to the variants described in \cite{camoriano2016nytro}. The graph shows the test error on the HIGGS dataset ($1.1 \times 10^7$ examples) with respect to the number of iterations (epochs for stochastic algorithms). \label{fig:falkon-vs-others}}
	\end{figure}
	
	We then recall the standard, albeit technical,  assumptions leading to fast  rates \cite{caponnetto2007optimal,steinwart2009optimal}. The {\em capacity condition} requires the existence of $\gamma \in (0,1]$ and $Q \geq 0$, such that ${\cal N}(\la) \leq Q^2 \la^{-\gamma}$. Note that this condition is always satisfied with $Q = \kappa$ and $\gamma = 1$. The {\em source condition} requires the existence of $r \in [1/2, 1]$ and $g \in \hh$, such that  $\fh = C^{r-1/2} g$. Intuitively, the {\em capacity condition} measures the size of $\hh$, if $\gamma$ is small then  $\hh$ is small and rates are faster. The {\em source condition} measures the regularity of $f_\hh$, if $r$ is big $f_\hh$ is regular and rates are faster. The case $r=1/2$ and $\gamma=D/(2s)$ (for a kernel with smoothness $s$ and input space $\R^D$) recovers the classic Sobolev condition. For further discussions on the interpretation of the conditions above see \cite{caponnetto2007optimal,steinwart2009optimal,conf/colt/Bach13,rudi2015less}.  We can then state our main result on fast rates
	\bt\label{thm:fast-rates}
	Let $\delta \in (0,1]$. 
	\loz{Assume  there exists $\kappa \geq 1$ such that $K(x,x) \leq \kappa^2$ for any $x \in \X$, and}
	$y \in [-\frac{a}{2}, \frac{a}{2}]$, almost surely, with $a > 0$. There exist an $n_0 \in \N$ such that for any $n \geq n_0$ the following holds. When
	$$\la = n^{-\frac{1}{2r + \gamma}}, \qquad \iter ~\geq~ \log(n) ~+~ 5 + 2\log (a + 3\kappa^2),$$
	${}\quad 1.$ and either \Nystrom{} uniform sampling is used with $M \geq 70\left[1 + {\cal N}_\infty(\la)\right]\log \frac{8\kappa^2}{\la \delta},$ \\
	${}\quad 2.$ or \Nystrom{} $q$-approx. lev. scores \cite{rudi2015less} is used with $M \geq 220\left[2 + q^2{\cal N}(\la)\right]\log \frac{8\kappa^2}{\la \delta},$
	
	\noindent then with probability $1-\delta$,
	$$ {\cal R}(\widehat{f}_{\la, M, t}) ~\leq~ c_0 \log^2 \frac{24}{\delta}~~n^{-\frac{2r}{2r+\gamma}}.$$
	where $\widehat{f}_{\la, M, t}$ is the FALKON estimator (Sect.~\ref{sect:FALKON}, Alg.~\ref{algo:main} and Sect.~\ref{sect:gen-algo}, Alg.~\ref{algo:main-MATLAB} in the appendix for the complete version). In particular $n_0, c_0$ do not depend on $\la, M, n, t$ and $c_0$ do not depend on $\delta$.
	\et
	The above result shows that FALKON achieves the same fast rates as KRR, under the same conditions \cite{caponnetto2007optimal}. For $r=1/2, \gamma=1$, the rate in Thm.~\ref{thm:simple-rates} is recovered. If $\gamma < 1, r > 1/2$, FALKON achieves a rate close to $O(1/n)$. By selecting the \Nystrom{} points with uniform sampling, a bigger $M$ could be needed for fast rates (albeit always less than $n$). However,  when approximate leverage scores are used $M$, smaller than $n^{\gamma/2} \ll \sqrt{n}$ is always enough for optimal generalization. \loz{This shows that FALKON with approximate leverage scores is the first algorithm  to achieve fast rates with a computational complexity that is $O(n {\cal N}(\la)) = O(n^{1 + \frac{\gamma}{2r+\gamma}}) \leq O(n^{1+\frac{\gamma}{2}})$ in time.}
	%
	%
	%
	%
	%
	%
	\section{Experiments}\label{sect:experiments}
\begin{table}[t]
	\caption{Architectures: $\ddagger$ cluster 128 EC2 r3.2xlarge machines,
		$\dagger$ cluster 8 EC2 r3.8xlarge machines, $\wr$ single machine with two Intel Xeon E5-2620,
		one Nvidia GTX Titan X GPU, 128GB RAM, $\star$ cluster with IBM POWER8 12-core processor, 512 GB RAM, $\ast$ unknown platform.}
	\label{MYT_table}
	\centering
	\noindent\resizebox{\textwidth}{!}{%
		\begin{tabular}{@{}lccccccc@{}}
			\toprule
			& \multicolumn{3}{c}{MillionSongs} & \multicolumn{2}{c}{YELP} & \multicolumn{2}{c}{TIMIT}    \\
			\cmidrule(lr){2-4} \cmidrule(lr){5-6} \cmidrule(lr){7-8}
			& MSE & Relative error & Time(\textit{s}) & RMSE & Time(\textit{m}) & c-err & Time(\textit{h}) \\
			\midrule
			FALKON                      & \textbf{80.10}  &$\mathbf{4.51\times10^{-3}}$    & \textbf{55} & \textbf{0.833} & \textbf{20} & 32.3\% & \textbf{1.5}\\
			Prec. KRR \cite{avron2016faster}                 &-               &$4.58\times10^{-3}$    &$289^\dagger$ &- &- &- &- \\
			Hierarchical \cite{Chen2016HierarchicallyCK}                 &-               &$4.56\times10^{-3}$    &$293^\star$ &- &- &- &-\\
			D\&C \cite{conf/colt/ZhangDW13}            & 80.35    &-             & $737^\ast$ &- &- &- &-\\
			Rand. Feat. \cite{conf/colt/ZhangDW13}     & 80.93    &-            & $772^\ast$ &- &- &- &-\\
			\Nystrom{} \cite{conf/colt/ZhangDW13}             & 80.38    &-            & $876^\ast$ &- &- &- &-\\
			ADMM R. F.\cite{avron2016faster}                 &-               &$5.01\times10^{-3}$    &$958^\dagger$ &- &- &- &-\\
			BCD R. F. \cite{tu2016large}       &- &- &-                & 0.949             & $42^\ddagger$ & 34.0\% & $1.7^\ddagger$\\
			BCD \Nystrom{} \cite{tu2016large}       &- &- &-                & 0.861             & $60^\ddagger$ & 33.7\% & $1.7^\ddagger$ \\
			EigenPro \cite{ma2017diving}    &- &- &- &- &-        & 32.6\% & $3.9^\wr$ \\
			KRR  \cite{Chen2016HierarchicallyCK} \cite{tu2016large}          &- &$4.55\times10^{-3}$ &- & 0.854             & $500^\ddagger$ & 33.5\% & $8.3^\ddagger$ \\
			Deep NN \cite{May2017KernelAM}  &- &- &- &- &-                 & 32.4\% & -   \\
			Sparse Kernels \cite{May2017KernelAM} &- &- &- &- &-   & \textbf{30.9\%} & -   \\
			Ensemble \cite{Huang2014KernelMM}&- &- &- &- &-          & 33.5\% & -   \\
			\bottomrule
		\end{tabular}
	}
\end{table}
We present FALKON's performance on a range of large scale datasets.  \loz{ As shown in Table~\ref{MYT_table},~\ref{SH_table}, FALKON achieves state of the art accuracy and typically outperforms previous approaches in all the considered large scale datasets including IMAGENET. This is remarkable considering  FALKON required only  a fraction of the competitor's computational resources. 
Indeed we used a single machine equipped with two Intel Xeon
	E5-2630 v3, one NVIDIA
	Tesla K40c and 128 GB of RAM and a basic MATLAB FALKON implementation, while typically the results for competing algorithm have been performed on clusters of GPU workstations (accuracies, times and used architectures are cited from the corresponding papers).}

A minimal MATLAB implementation of FALKON is presented in Appendix~\ref{sect:matlab}. The code necessary to reproduce the following experiments, plus a FALKON version that is able to use the GPU, is available on GitHub at 
\url{https://github.com/LCSL/FALKON_paper}

The error is measured with MSE, RMSE or relative error for regression problems,
and with classification error (c-err) or AUC for the classification problems, to be consistent with the literature.
For datasets which do not have a fixed test set,
we set apart 20\% of the data for testing.
For all datasets, but YELP and IMAGENET, we normalize the features by their z-score.
From now on we denote with $n$ the cardinality of the dataset, $d$ the dimensionality. A comparison of FALKON  with respect to other methods to compute the \Nystrom{} estimator, in terms of the MSE test error on the HIGGS dataset, is given in Figure~\ref{fig:falkon-vs-others}.
\\ \\{\bf MillionSongs} \cite{BertinMahieux2011TheMS} (Table~\ref{MYT_table}, $n=4.6 \times 10^5$, $d=90$, regression).
We used a Gaussian kernel with $\sigma = 6$, $\lambda = 10^{-6}$ and $10^4$ \Nystrom{} centers.
Moreover with $5\times10^4$ center, FALKON achieves a $79.20$ MSE, and $4.49\times10^{-3}$ rel. error in $630$ \textit{sec}.
\\ \\{\bf TIMIT}~(Table~\ref{MYT_table}, $n = 1.2 \times 10^6$, $d = 440$, multiclass classification).
We used the same preprocessed dataset of \cite{ma2017diving} and Gaussian Kernel with $\sigma = 15$,
$\lambda = 10^{-9}$ and $10^5$ \Nystrom{} centers.
\\ \\{\bf YELP}~~(Table~\ref{MYT_table}, $n = 1.5 \times 10^6$, $d = 6.52 \times 10^7$, regression).
We used the same dataset of \cite{tu2016large}. We extracted the 3-grams from the plain text with the same pipeline as \cite{tu2016large},
then we mapped them in a sparse binary vector which records if the 3-gram is present or not in the example. We used a linear kernel with $5\times10^4$ \Nystrom{} centers. With $10^5$ centers, we get a RMSE of $0.828$ in $50$ minutes.
\begin{table}[t]
  \caption{Architectures: $\dagger$ cluster with IBM POWER8 12-core cpu, 512 GB RAM,
        $\wr$ single machine with two Intel Xeon E5-2620, one Nvidia GTX Titan X GPU, 128GB RAM,
        $\ddagger$ single machine \cite{Alves2016StackingML}}
   \label{SH_table}
  \centering
  \noindent\resizebox{\textwidth}{!}{%
  \begin{tabular}{@{}lcccccccc@{}}
    \toprule
        & \multicolumn{3}{c}{SUSY} & \multicolumn{2}{c}{HIGGS}& \multicolumn{2}{c}{IMAGENET}   \\
        \cmidrule(lr){2-4} \cmidrule(lr){5-6} \cmidrule(lr){7-8}
        & c-err & AUC & Time(\textit{m}) & AUC & Time(\textit{h}) & c-err & Time(\textit{h})\\
    \midrule
    FALKON                      & \textbf{19.6\%}  & 0.877   & \textbf{4} & 0.833 & \textbf{3} &20.7\% & \textbf{4}\\
    EigenPro \cite{ma2017diving}    & 19.8\%         &-       & $6^\wr$ &- &- &- &-\\
    Hierarchical \cite{Chen2016HierarchicallyCK}      & 20.1\%         &-       & $40^\dagger$  &- &-&- &-\\
    Boosted Decision Tree \cite{baldi2014searching}                     &-              & 0.863     &-  & 0.810 &-&- &-\\
    Neural Network \cite{baldi2014searching}                   &-                 & 0.875    & - & 0.816 & -&- &-\\
    Deep Neural Network \cite{baldi2014searching}                  &-                  &\textbf{0.879}     & $4680^\ddagger$   & \textbf{0.885} & $78^\ddagger$&- &-\\
    Inception-V4 \cite{iv42017}                   &-                 & -    & - & - & -&\textbf{20.0\%} &-\\
    \bottomrule
  \end{tabular}
  }
\end{table}
\\ \\{\bf SUSY}~(Table~\ref{SH_table}, $n = 5\times10^6$, $d = 18$, binary classification).
We used a Gaussian kernel with $\sigma = 4$, $\lambda = 10^{-6}$ and $10^4$ \Nystrom{} centers.
\\ \\{\bf HIGGS}~(Table~\ref{SH_table}, $n = 1.1\times 10^7$, $d = 28$, binary classification).
Each feature has been normalized subtracting its mean and dividing for its variance.
We used a Gaussian kernel with diagonal matrix width learned with cross validation on a small validation set,
$\lambda = 10^{-8}$ and
$10^5$ \Nystrom{} centers.
If we use a single $\sigma = 5$ we reach an AUC of $0.825$.
\\ \\{\bf IMAGENET}~(Table~\ref{SH_table}, $n = 1.3\times 10^6$, $d = 1536$, multiclass classification).
We report the top 1 c-err over the validation set of ILSVRC 2012 with a single crop. The features are obtained from the convolutional layers of pre-trained Inception-V4 \cite{iv42017}. 
We used Gaussian kernel with $\sigma = 19$, $\lambda = 10^{-9}$ and
$5\times 10^4$ \Nystrom{} centers. Note that with linear kernel we achieve c-err $= 22.2\%$.

		\paragraph{Acknowledgments.}
		The authors would like to thank Mikhail Belkin, Benjamin Recht and Siyuan Ma, Eric Fosler-Lussier, Shivaram Venkataraman, Stephen L. Tu, for providing their features of the TIMIT and YELP datasets, and NVIDIA Corporation for the donation of the Tesla K40c GPU used for this research.
		This work is funded by the Air Force project FA9550-17-1-0390 (European Office of Aerospace Research and Development) and by the FIRB project RBFR12M3AC (Italian Ministry of Education, University and Research).
		\bibliographystyle{unsrt}
		\bibliography{nipsBib}

	\newpage
	
	\appendix
	
	{\LARGE FALKON: An Optimal Large Scale Kernel Method}
	
	{\Large Supplementary Materials}\\
	
	\begin{enumerate}
		\item[\ref{sect:gen-algo}.] {\em FALKON: General Algorithm} \\where a generalized version of FALKON able to deal with both invertible and non-invertible $\Km$ is provided.
		
		\item[\ref{sect:notation}.] {\em Definitions and Notation} \\where the notation, required by the proofs, is given and the basic operators are defined.
		\item[\ref{sect:analytic-dec}.] {\em Analytic Decompositions} \\where the condition number of the FALKON system is controlled and the excess risk of FALKON is decomposed in terms of functional analytic quantities.
		\item[\ref{sect:prob-est}.] {\em Probabilistic Estimates} \\where the quantities of the previous section are bounded in probability.
		\item[\ref{sect:proof-main-res}.] {\em Proof of the Main Results} \\where the results of the previous sections are collected and the proofs of the main theorems of the paper are provided
		\item[\ref{sect:long-comparison}.] {\em Longer Comparison with the Literature} \\where some more details on previous works on the topic are given.
		\item[\ref{sect:matlab}.] {\em MATLAB Code for FALKON} \\where a minimal working implementation of FALKON is provided.
	\end{enumerate}

	\small

	\section{FALKON: General Algorithm}\label{sect:gen-algo}
	In this section we define a generalized version of FALKON. In particular we provide a preconditioner able to deal with non invertible $\Km$ and with \Nystrom{} centers selected by using approximate leverage scores. In Def.~\ref{def:matr-algo} we state the properties that such preconditioner must satisfy. In Example~\ref{ex:gen-prec-falcon} we show that the preconditioner in Sect.~\ref{sect:FALKON} satisfies Def.~\ref{def:matr-algo} when $\Km$.
	
	First we recall some ways to sample \Nystrom{} centers, from the training set.
	
	{\bf \Nystrom{} with uniform sampling.} Let $n, M \in \N$ with $1 \leq M \leq n$. Let $x_1,\dots, x_n$ be the training set. The \Nystrom{} centers $\widetilde{x}_1, \dots, \widetilde{x}_M$ are a random subset of cardinality $M$ uniformly sampled from the training set.
	
	{\bf \Nystrom{} with approximate leverage scores.} We recall the definition of approximate leverage scores, from \cite{rudi2015less} and then the sampling method based on them. Let $n \in \N$, $\la > 0$. Let $x_1,\dots, x_n$ be the training points and define $\K \in \R^{n\times n}$ as $(\K)_{ij} = K(x_i, x_j)$ for $1 \leq i,j \leq n$. The exact leverage scores are defined by
	$$ l_\la(i) = \left(\K(\K + \la n I)^{-1}\right)_{ii},$$
	for any $i \in {1,\dots, n}$. Any bi-lipschitz approximation of the exact leverage scores, satisfying the following definition is denoted as approximate leverage scores.
	\bd[\Nystrom{} with $(q,\la_0, \delta)$-approximate leverage scores \cite{rudi2015less}]\label{def:emp-q-als}
	Let $\delta \in (0,1]$ and $\la_0 > 0$ and $q \in [1, \infty)$. A (random) sequence $(\widehat{l}_\la(i))_{i=1}^n$ is denoted as $(q, \la_0, \delta)$-approximate leverage scores when the following holds with probability at least $1-\delta$
	$$ \frac{1}{q} ~l_\la(i) ~\leq ~\widehat{l}_\la(i)~ \leq ~ q ~ l_\la(i), \qquad \forall \la \geq \la_0, t \in \{1, \dots, n\}.$$
	\ed
	In particular, given $n \in \N$ training points $x_1,\dots, x_n$, and a sequence of approximate leverage scores $(\widehat{l}_\la(i))_{i=1}^n$, the \Nystrom{} centers are selected in the following way. Let $p_i = \frac{\widehat{l}_\la(i)}{\sum_{j=1}^n \widehat{l}_\la(j)}$, with $1 \leq i \leq n$. Let $i_1, \dots, i_M$ be independently sampled from $\{1,\dots, n\}$ with probability $(p_i)_{i=1}^n$. Then $\widetilde{x}_1 := x_{i_1},\dots, \widetilde{x}_M := x_{i_M}$.

	Now we define a diagonal matrix depending on the used sampling scheme that will be needed for the general preconditioner.
	\bd\label{def:diag-D}
	Let $D \in \R^{M\times M}$ be a diagonal matrix. If the \Nystrom{} centers are selected via uniform sampling, then $D_{jj} = 1$, for $1 \leq j \leq M$. \\
	Otherwhise, let $i_1,\dots,i_M \in \{1,\dots, n\}$ be the indexes of the training points sampled via approximate leverage scores. Then for $1 \leq j \leq M$,
	$$D_{jj} = \sqrt{\frac{1}{n p_{i_j}}}.$$
	\ed
	We note here that by definition $D$ is a diagonal matrix with strictly positive and finite diagonal. Indeed it is true in the uniform case. In the leverage scores case, let $1 \leq j \leq M$.  Note that since the index $i_j$ has been sampled, it implies that the probability $p_{i_j}$ is strictly larger than zero. Then, since $0 < p_{i_j} \leq 1$ then $0 < D_{jj} < \infty$ a.s.~. 
	
	\subsection{Generalized FALKON Algorithm} 
	We now introduce some matrices needed for the definition of a generalized version of FALKON, able to deal with non invertible $\Km$ and with different sampling schemes, for the \Nystrom{} centers. Finally in Def.~\ref{def:generalized-FALKON}, we define a general form of the algorithm, that will be used in the rest of the appendix.
	
	\bd[The generalized preconditioner]\label{def:matr-algo}
	Let $M \in \N$. Let $\widetilde{x}_1, \dots, \widetilde{x}_M \in \X$ and $\Km \in \R^{M\times M}$ with $(\Km)_{ij} = K(\widetilde{x}_i,\widetilde{x}_j)$, for $1\leq i,j \leq M$. Let $D \in \R^{M\times M}$ be a diagonal matrix with strictly positive diagonal, defined according to Def~\ref{def:diag-D}.
	
	Let $\la > 0$, $q \leq M$ be the rank of $\Km$, $Q \in \R^{M \times q}$ a partial isometry such that $Q^\top Q = I$ and $T \in \R^{q\times q}$ a triangular matrix. Moreover $Q, T$ satisfy the following equation
	$$ D \Km D = Q T^\top T Q^\top .$$
	Finally let $A \in \R^{q \times q}$ be a triangular matrix such that
	$$ A^\top A = \frac{1}{M} T T^\top + \lambda I.$$
	Then the {\em generalized preconditioner} is defined as
	$$B ~= ~\frac{1}{\sqrt{n}}~ D Q T^{-1} A^{-1}.$$
	\ed
	Note that $B$ is right invertible, indeed $D$ is invertible, since is a diagonal matrix, with strictly positive diagonal, $T, A$ are invertible since they are square and full rank and $Q$ is a partial isometry, so $B^{-1} = \sqrt{n} A T Q^{\top} D^{-1}$ and $B B^{-1}  = I$. Now we provide two ways to compute $Q, T, A$. We recall that the Cholesky algorithm, denoted by $\rm chol$, given a square positive definite matrix, $B \in \R^{M \times M}$, produces an upper triangular matrix $R \in \R^{M \times M}$ such that $B = R^\top R$. While the pivoted (or rank revealing) QR decomposition, denoted by $\rm qr$, given a square matrix $B$, with rank $q$, produces a partial isometry $Q \in \R^{M \times q}$ with the same range of $M$ and an upper trapezoidal matrix $R \in \R^{q \times M}$ such that $B = QR$.
	\bex[precoditioner satisfying Def.~\ref{def:matr-algo}]\label{ex:gen-prec-falcon}
	Let $\la > 0$, and $\Km, D$ as in Def.~\ref{def:matr-algo}. 
	\begin{enumerate}
		\item When $\Km$ is full rank ($q = M$), then the following $Q, T, A$ satisfy Def.~\ref{def:matr-algo}
		$$ Q = I, \quad T = {\rm chol}(D\Km D), \quad A = {\rm chol}\left(\frac{1}{M}TT^\top + \lambda I\right).$$
		\item When $\Km$ is of any rank ($q \leq M$), then the following $Q, T, A$ satisfy Def.~\ref{def:matr-algo}
		$$ (Q, R) = {\rm qr}(D \Km D), \quad T = {\rm chol}(Q^\top D\Km D Q), \quad A = {\rm chol}\left(\frac{1}{M}TT^\top + \lambda I\right).$$
	\end{enumerate}
	\eex
	\bpr
	In the first case, $Q, T, A$ satisfy Def.~\ref{def:matr-algo} by construction.
	In the second case, since $QQ^\top$ is the projection matrix on the range of $D\Km D$, then $QQ^\top D\Km D = D\Km D$ and, since $D\Km D$ is symmetric, $D \Km D QQ^\top = D\Km D$, so
	$$QT^\top T Q^\top = Q Q^\top D \Km D Q Q^\top = D \Km D.$$
	Moreover note that, since the rank of $\Km$ is $q$, then the range of $D \Km D$ is $q$, and so $Q^\top Q = I$, since it is a partial isometry with dimension $\R^{M \times q}$. Finally $A$ satisfies Def.~\ref{def:matr-algo} by construction.
	\epr
	
	Instead of rank-revealing QR decomposition, eigen-decomposition can be used.
	\bex[preconditioner for the deficient rank case, using $\rm eig$ instead of $\rm qr$]
	Let $\la > 0$, and $\Km, D$ as in Def.~\ref{def:matr-algo}. Let $(\la_i, u_i)_{1 \leq i \leq M}$ be respectively the eigenvalues and the associated eigenvectors from the eigendecomposition of $D\Km D$, with $\la_1 \geq \dots \geq \la_M \geq 0$. So the following $Q,T,A$ satisfy Def.~\ref{def:matr-algo}, $Q = (u_1,\dots, u_q)$ and $T = {\rm diag}(\sqrt{\la_1},\dots,\sqrt{\la_q})$, while $A = {\rm diag}\left(\sqrt{\la + \frac{1}{M}\la_1},\dots,\sqrt{\la + \frac{1}{M}\la_q}\right)$. 
	\eex
	We recall that this approach to compute $Q,T,A$ is conceptually simpler than the one with QR decomposition, but slower, since the hidden constants in the eigendecomposition are larger than the one of QR.
	
	The following is the general form of the algorithm.
	\bd[Generalized FALKON algorithm]\label{def:generalized-FALKON}
	Let $\la > 0$, $\iter \in \N$ and $q, Q, T, A$ as in Def.~\ref{def:matr-algo}. The generalized FALKON estimator is defined as follows
	$$\widehat{f}_{\la, M, t}(x) = \sum_{i=1}^M \alpha_i K(x, \tilde{x}_i), \quad \textrm{with} \quad \alpha = B \beta_{\iter},$$
	and $\beta_\iter \in \R^q$ denotes the vector resulting from $\iter$ iterations of the conjugate gradient algorithm applied to the following linear system
	\eqal{\label{eq:gen-FALKON-problem}
		W \beta = b, \quad \textrm{where} \quad W= B^\top (\Knm^\top \Knm + \lambda n \Km) B, \quad b = B^\top \Knm^\top \yn.
	}
	\ed
	
	\section{Definitions and Notation} \label{sect:notation}
	Here we recall some basic facts on linear operators and give some notation that will be used in the rest of the appendix, then we define the necessary operators to deal with the excess risk of FALKON via functional analytic tools.

	\paragraph{Notation}
	Let $\hh$ be an Hilbert space, we denote with $\nor{\cdot}_\hh$, the associated norm and with $\scal{\cdot}{\cdot}_\hh$ the associated inner product.
	We denote with $\nor{\cdot}$ the operator norm for a bounded linear operator $A$, defined as
	$\nor{A} = \sup_{\nor{f}_\hh = 1} \nor{A f}$. Moreover we will denote with $\otimes$ the tensor product, in particular
	$$(u \otimes v)z = u \scal{v}{z}_\hh, \quad \forall u, v, z \in \hh.$$
	In the rest of the appendix $A + \la I$ is often denoted by $A_\la$ where $A$ is linear operator and $\la \in \R$, moreover we denote with $A^*$ the adjoint of the linear operator $A$, we will use $A^\top$ if $A$ is a matrix.
	When $\hh$ is separable, we denote with $\tr$ the trace, that is $\tr(A) = \sum_{j=1}^D \scal{u_i}{A u_i}_\hh$ for any linear operator $A:\hh \to \hh$, where $(u_i)_{j=1}^D$ is an orthogonal basis for $\hh$ and $D \in \N \cup \{\infty\}$ is the dimensionality of $\hh$. Moreover we denote with $\nor{\cdot}_\HS$ the Hilbert-Schmidt norm, that is
	$\nor{A}_\HS^2 = \tr(A^*A)$, for a linear operator $A$.
	
	In the next proposition we recall the spectral theorem for compact self-adjoint operators on a Hilbert space.
	\bp[Spectral Theorem for compact self-adjoint operators]\label{prop:spectral-theorem-compact}
	Let $A$ be a compact self-adjoint operator on a separable Hilbert space $\hh$. Then there exists a sequence $(\la_j)_{j=1}^D$ with $\la_j \in \R$, and an orthogonal basis of $\hh$ $(u_j)_{j=1}^D$ where $D \in \N \cup \{\infty\}$ is the dimensionality of $\hh$, such that
	\eqal{\label{eq:spectral-theorem}
		A = \sum_{j=1}^D \la_j u_j \otimes u_j.
	}
	\ep
	\bpr
	Thm. VI.16, pag. 203 of \cite{reed1980methods}.
	\epr
	Let $\hh$ be a separable Hilbert space (for the sake of simplicity assume $D = \infty$), and $A$ be a bounded self-adjoint operator on $\hh$ that admits a spectral decomposition as in Eq.~\ref{eq:spectral-theorem}. Then the largest and the smallest eigenvalues of $A$ are denoted by
	\eqals{
		\la_{\max}(A) = \sup_{j \geq 1} \la_j, \quad \la_{\min}(A) = \inf_{j \geq 1} \la_j.
	}
	In the next proposition we recall a basic fact about bounded symmetric linear operators on a separable Hilbert space $\hh$.
	\bp\label{prop:spectral-calculus}
	Let $A$ be a bounded self-adjoint operator on $\hh$, that admits a spectral decomposition as in Eq.~\ref{eq:spectral-theorem}. Then
	$$-\nor{A} \leq \la_{\min}(A) \leq \la_{\max}(A) \leq \nor{A}.$$
	\ep
	\bpr
	By definition of operator norm, we have that $\nor{Ax}^2_\hh \leq \nor{A}^2 \nor{x}^2_\hh \quad \forall  x \in \hh$.
	Let $(\la_j, u_j)_{j=1}^D$ be an eigendecomposition of $A$, with $D$ the dimensionality of $\hh$, according to Prop.~\ref{prop:spectral-theorem-compact}, then, for any $j \geq 1$, we have
	$$ \la_j^2 = \scal{A u_j}{A u_j} =  \nor{A u_j}_\hh^2 \leq \nor{A}^2,$$
	where we used the fact that $A u_j = \la_j u_j$ and that $\nor{u_j}_\hh = 1$.
	\epr
	%
	%
	%
	%
	%
	\subsection{Definitions}
	Let $\X$ be a measurable and separable space and $\Y = \R$. Let $\rho$ be a probability measure on $\X \times \R$. We denote with $\rhox$ the marginal probability of $\rho$ on $\X$ and with $\rho(y|x)$ the conditional probability measure on $\Y$ given $\X$. Let $\Ltwo$ be the Lebesgue space of $\rhox$ square integrable functions, endowed with the inner product
	\eqals{
		\scal{\phi}{\psi}_\rho = \int \phi(x)\psi(x) d\rhox(x), \quad \forall \phi, \psi \in \Ltwo,
	}
	and norm $\nor{\psi}_\rho = \sqrt{\scal{\psi}{\psi}}_\rho$ for any $\psi \in \Ltwo$. We now introduce the kernel and its associated space of functions. Let $K: \X \times \X \to \R$ be a positive definite kernel, measurable and uniformly bounded, i.e. there exists $\kappa \in (0, \infty)$, for which $K(x,x) \leq \kappa^2$ almost surely. We denote with $K_x$ the function $K(x, \cdot)$ and with $(\hh,~\scal{\cdot}{\cdot}_\hh)$, the Hilbert space of functions with the associated inner product induced by $K$, defined by
	\eqals{
		\hh = \lspanc{K_x}{x \in \X}, \quad \scal{K_x}{K_{x'}}_\hh = K(x,x'), ~~ \forall ~ x,x' \in \X.
	}

	Now we define the linear operators used in the rest of the appendix
	\bd\label{def:ideal-op}
	Under the assumptions above, for any $f \in \hh, \phi \in \Ltwo$
	\begin{itemize}
		\item $S: \hh \to \Ltwo$, such that $Sf = \scal{f}{K_{(\cdot)}}_{\hh} \in \Ltwo$, with adjoint
		\item $S^*: \Ltwo \to \hh$, such that $S^*\phi = \int \phi(x) K_x d\rhox(x) \in \hh$.
		\item $L: \Ltwo \to \Ltwo$, such that $L = SS^*$ and
		\item $C: \hh \to \hh$, such that $C = S^*S$.
	\end{itemize}
	\ed
	Let $x_i \in \X$ with $1 \leq i \leq n$ and $n \in \N$, and $\widetilde{x}_j \in \X$ for $1 \leq j \leq M$ and $M \in \N$. We define the following linear operators
	\bd\label{def:emp-op}
	Under the assumptions above, for any $f \in \hh, v \in \R^n, w \in \R^M$,
	\begin{itemize}
		\item $\Sn: \hh \to \R^n$, such that $\Sn f = \frac{1}{\sqrt{n}}(\scal{f}{K_{x_i}})_{i=1}^n \in \R^n$, with adjoint
		\item $\Sn^*: \R^n \to \hh$, such that $\Sn^* v = \frac{1}{\sqrt{n}}\sum_{i=1}^n v_i K_{x_i} \in \hh.$
		\item $\Cn: \hh \to \hh$, such that $\Cn = \Sn^*\Sn$.
		\item $\Sm: \hh \to \R^M$, such that $\Sm f = \frac{1}{\sqrt{M}}(\scal{f}{K_{\tilde{x}_i}})_{i=1}^M \in \R^M$, with adjoint
		\item $\Sm^*: \R^M \to \hh$, such that $\Sm^* w = \frac{1}{\sqrt{M}}\sum_{i=1}^M v_i K_{\tilde{x}_i} \in \hh.$
		\item $\Cm: \hh \to \hh$, such that $\Cm = \Sm^*\Sm$.
		\item $\Gm: \hh \to \hh$, such that $\Gm = \Sm^*D^2\Sm$, with $D$ defined in Def.~\ref{def:matr-algo} (see also Def.~\ref{def:diag-D}).
	\end{itemize}
	\ed

	We now recall some basic facts about $L, C, S, K_{nn}, \Cn, \Sn$, $\Knm$ and $\Km$.
	\bp\label{prop:basic_operator_result}
	With the notation introduced above,
	\eqals{
		1.& & \Knm & = \sqrt{nM} \Sn\Sm^*, & \quad \Km = M ~ \Sm\Sm^*,  \qquad\qquad \K = n ~ \Sn\Sn^* \quad\quad\qquad &\\
		2.& & C &= \int_\X K_x \otimes K_x d\rhox(x), & \tr(C) =  \tr(L) = \|S\|_{HS}^2 = \int_\X \|K_x\|_\hh^2 d\rhox(x) &\leq \kappa^2, \\
		3.& &\Cn& = \frac{1}{n}\sum_{i=1}^n K_{x_i} \otimes K_{x_i}, & \tr(\Cn) = \tr(K_{nn}/n)  = \|\Sn\|_{HS}^2 = \frac{1}{n}\sum_{i=1}^n \nor{K_{x_i}}^2_\hh &\leq \kappa^2, \\
		4.& &\Cm& = \frac{1}{M}\sum_{i=1}^M K_{\tilde{x}_i} \otimes K_{\tilde{x}_i}, & \tr(\Cm) = \tr(\Km/M)  = \|\Sm\|_{HS}^2 = \frac{1}{M}\sum_{i=1}^m \nor{K_{\tilde{x}_i}}^2_\hh &\leq \kappa^2, \\
		5.& &\Gm& = \frac{1}{M}\sum_{i=1}^M D^2_{ii} K_{\tilde{x}_i} \otimes K_{\tilde{x}_i}.
	}
	where $\otimes$ denotes the tensor product.
	\ep
	\bpr
	Note that $(\Knm)_{ij} = K(x_i,\tilde{x}_j) = \scal{K_{x_i}}{K_{\tilde{x}_j}}_\hh = (\sqrt{nM} \Sn\Sm^*)_{ij}$, for any $1 \leq i \leq n$, $1 \leq j \leq M$, thus
	$\Knm = \sqrt{nM} \Sn\Sm^*$. The same reasoning holds for $\Km$ and $\K$.
	For the second equation, by definition of $C = S^*S$ we have that, for each $h,h'\in\hh$,
	\eqals{
		\scal{h}{Ch'}_\hh &= \scal{Sh}{Sh'}_\rho = \int_\X \scal{h}{K_x}_\hh \scal{K_x}{h'}_\hh d\rhox(x)  = \int_\X \scal{h}{\Big(K_x\scal{K_x}{h'}_\hh\Big)}_\hh d\rhox(x) \\
		& = \int_\X \scal{h}{\Big(K_x\otimes K_x\Big)h'}_\hh d\rhox(x) = \scal{h}{\Big(\int_\X K_x\otimes K_x d\rhox(x)\Big)h'}_\hh.
	}
	Note that, since $K$ is bounded almost surely, then $\|K_x\|_\hh \leq \kappa$ for any $x \in \X$, thus
	\eqals{
		\tr(C) = \int_\X \tr(K_x \otimes K_x) d\rhox(x) = \int_\X \|K_x\|_\hh^2 d\rhox(x) \leq \kappa^2
	}
	by linearity of the trace. Thus $\tr(C) < \infty$ and so
	$$\tr(C) = \tr(S^*S) = \|S\|_{HS}^2 = \tr(S S^*) = \tr(L).$$
	The proof for the rest of equations is analogous to the one for the second.
	\epr
	Now we recall a standard characterization of the excess risk
	\bp\label{prop:gen-error}
	When $\int_\Y y^2 d\rho < \infty$, then there exist $\frho \in \Ltwo$ defined by
	$$ \frho(x) = \int y d\rho(y|x),$$
	almost everywhere. Moreover, for any $\hat{f} \in \hh$ we have,
	$${\cal E}(\widehat{f}~) - \inf_{f \in \hh}{\cal E}(f) = \nor{S\widehat{f} - P\frho}_\rhox^2,$$
	where $P: \Ltwo \to \Ltwo$ is the projection operator whose range is the closure in $\Ltwo$ of the range of $S$.
	\ep
	\bpr
	Page 890 of \cite{vito2005learning}.
	\epr

	%
	
	\section{Analytic results}\label{sect:analytic-dec}
	The section of analytic results is divided in two subsections, where we bound the condition number of the FALKON preconditioned linear system \eqref{eq:gen-FALKON-problem} and we decompose the excess risk of FALKON, with respect to analytical quantities that will be controlled in probability in the following sections.
	
	\subsection{Analytic results (I): Controlling condition number of $W$}
	First we characterize the matrix $W$ defining the FALKON preconditioned linear system \eqref{eq:gen-FALKON-problem}, with respect to the operators defined in Def.~\ref{def:emp-op} (see next lemma) and in particular we characterize its condition number with respect to the norm of an auxiliary  operator defined in Lemma~\ref{lm:cond-control}. Finally we bound the norm of such operator with respect to analytical quantities more amenable to be bounded in probability (Lemma~\ref{lm:control-E}).
	\blm[Characterization of $W$]\label{lm:characterization-W}
	Let $\la \in \R$. The matrix $W$ in Def.~\ref{def:generalized-FALKON} is characterized by
	$$ \quad W = A^{-\top}V^*(\Cn + \la I)VA^{-1}, \quad \textrm{with} \quad V = \sqrt{nM} \Sm^* B A.$$
	Moreover $V$ is a partial isometry such that $V^*V = I_{q \times q}$ and $V V^*$ with the same range of $\Sm^*$.
	\elm
	\bpr
	By the characterization of $\Knm, \Km$ and $\Cn$ in Prop.~\ref{prop:basic_operator_result}, we have
	\eqals{
		\Knm^\top \Knm + \la \Km &= nM ~ (\Sm\Sn^*\Sn\Sm^* + \la \Sm\Sm^*)  \\
		&=  nM ~ \Sm(\Sn^*\Sn + \la I)\Sm^*  = nM ~ \Sm(\Cn + \la I)\Sm^*.
	}
	Now note that, by definition of $B$ in Def.~\ref{def:matr-algo} and of $V$, we have
	$$\sqrt{nM} \Sm^* B = \sqrt{nM}\Sm^* B A A^{-1} = V A^{-1},$$
	so
	$$ W = B^\top (\Knm^\top \Knm + \la \Km) B ~= ~ n M ~~ B^\top\Sm(\Cn + \la I)\Sm^* B = A^{-\top} V^*(\Cn + \la I) V A^{-1}.$$
	The last step is to prove that $V$ is a partial isometry.
	First we need a characterization of $V$ that is obtained by expanding the definition of $B$,
	\eqal{\label{eq:char-V}
		V = \sqrt{nM} \Sm^* B A = \sqrt{nM} \Sm^* \frac{1}{\sqrt{n}} D Q T^{-1} A^{-1} A = \sqrt{M} \Sm^* D Q T^{-1}.
	}
	By the characterization of $V$, the characterization of $\Km$ in Prop.~\ref{prop:basic_operator_result} and the definition of $Q, T$ in terms of $D\Km D$ in Def.~\ref{def:matr-algo} , we have
	$$V^*V = M T^{-\top}Q^\top D~\Sm\Sm^*~ D Q T^{-1} = T^{-\top}Q^\top ~D\Km D~ Q T^{-1} = T^{-\top}Q^\top Q T^\top T Q^\top Q T^{-1} = I.$$
	Moreover, by the characterization of $V$, of $D\Km D$ with respect to $\Sm$, and of $Q, T$ (Prop.~\ref{prop:basic_operator_result} and Def.~\ref{def:matr-algo}),
	\eqals{
		VV^*\Sm^*D &= M ~ \Sm^*DQ T^{-1} T^{-\top} Q^\top D\Sm\Sm^* = \Sm^* D Q T^{-1} T^{-\top} Q^\top D \Km D  \\
		& = \Sm^* D Q T^{-1} T^{-\top} Q^\top Q T^\top T Q^\top = \Sm^* D Q Q^\top = \Sm^* D,
	}
	where the last step is due to the fact that the range of $QQ^\top$ is the one of $D\Km D$ by definition (see Def.~\ref{def:matr-algo}), and since $D\Km D = M D\Sm\Sm^* D$ by Prop.~\ref{prop:basic_operator_result}, it is the same of $D \Sm$. Note finally that the range of $\Sm^*D$ is the same of $\Sm^*$ since $D$ is a diagonal matrix with strictly positive elements on the diagonal (see Def.~\ref{def:matr-algo}).
	\epr
	
	\blm\label{lm:cond-control}
	Let $\la > 0$ and $W$ be as in Eq.~\ref{eq:gen-FALKON-problem}. Let $E = A^{-\top} V^*(\Cn - \Gm)VA^{-1}$, with $V$ defined in Lemma~\ref{lm:characterization-W}. Then $W$ is characterized by
	$$W = I + E.$$
	In particular, when $\nor{E} < 1$,
	\eqals{
		\cond{W} \leq \frac{1 + \nor{E}}{1 - \nor{E}}.
	}
	\elm
	\bpr
	Let $Q, T, A, D$ as in Def.~\ref{def:matr-algo}, and $V$ as in Lemma~\ref{lm:characterization-W}. According to Lemma~\ref{lm:characterization-W} we have
	$$W = A^{-\top} V^* (\Cn + \la I) V A^{-1} = A^{-\top} (V^*\Cn V + \la I) A^{-1}.$$
	Now we bound the largest and the smallest eigenvalue of $W$. First of all note that
	\eqal{\label{eq:dec-precond}
		A^{-\top} (V^*\Cn V + \la I) A^{-1} = A^{-\top} (V^*\Gm V + \la I) A^{-1} + A^{-\top} V^*(\Cn - \Gm) V A^{-1},
	}
	where $\Gm$ is defined in Def.~\ref{def:emp-op}.
	To study the first term, we need a preliminary result, which simplifies $\Sm V$. By using the definition of $V$, the characterization of $\Km$ in terms of $\Sm$ (Prop.~\ref{prop:basic_operator_result}), the definition of $B$ (Def.~\ref{def:matr-algo}), and finally the characterization of $D\Km D$ in terms of $Q, T$ (Def.~\ref{def:matr-algo}), we have
	\eqals{
		D \Sm V &= \sqrt{nM}D\Sm \Sm^* B A = \sqrt{\frac{n}{M}}D\Km B A = \frac{1}{\sqrt{M}} D\Km D~Q T^{-1} \\
		&= \frac{1}{\sqrt{M}}Q T^{\top}T Q^\top Q T^{-1} = \frac{1}{\sqrt{M}}Q T^{\top}.
	}
	Now we can simplify the first term. We express $\Gm$ with respect to $\Sm$, then we apply the identity above on $D \Sm V$ and on its transpose, finally we recall the identity $A^\top A = \frac{1}{M} TT^\top + \la I$ from Def.~\ref{def:matr-algo}, obtaining
	\eqal{\label{eq:AVCm-identity}
		A^{-\top} (V^*\Gm V + \la I) A^{-1} &= A^{-\top} (V^*\Sm^* D^2 \Sm V + \la I) A^{-1} = A^{-\top} ( \frac{1}{M}T Q^\top Q T^\top + \la I) A^{-1} \\
		& = A^{-\top} ( \frac{1}{M}T T^\top + \la I) A^{-1} = A^{-\top} A^\top A A^{-1} = I.
	}
	
	So, by defining $E := A^{-\top} V^*(\Cn - \Gm) V A^{-1}$, we have 
	$$W = I + E.$$
	Note that $E$ is compact and self-adjoint, by definition. Then, by Prop.~\ref{prop:spectral-theorem-compact},~\ref{prop:spectral-calculus} we have that $W$ admits a spectral decomposition as in Eq.~\ref{eq:spectral-theorem}.
	Let $\lambda_{\max}(W)$ and $\lambda_{\min}(W)$ be respectively the largest and the smallest eigenvalues of $W$, by Prop.~\ref{prop:spectral-calculus}, and considering that $-\nor{E} \leq \la_j(E) \leq \nor{E}$ (see Prop.~\ref{prop:spectral-theorem-compact}) we have
	\eqals{
		\lambda_{\max}(W) &=~ \sup_{j \in \N} 1 + \la_j(E)  ~=~ 1 ~+~ \sup_{j \in \N} \la_j(E) ~=~ 1 ~+~ \lambda_{\max}(E) ~\leq~ 1 + \nor{E},\\
		\lambda_{\min}(W) &=~ \inf_{j \in \N} 1 + \la_j(E)  ~=~ 1 ~+~ \inf_{j \in \N} \la_j(E) ~=~ 1 ~+~ \lambda_{\min}(E) ~\geq~ 1 - \nor{E}.
	}
	Since $W$ is self-adjoint and positive, when $\nor{E} < 1$, by definition of condition number, we have
	$$\cond{W} = \frac{\la_{\max}(W)}{\la_{\min}(W)} \leq \frac{1 + \nor{E}}{1 - \nor{E}}.$$
	\epr
	\blm\label{lm:control-E}
	Let $E$ be defined as in Lemma~\ref{lm:cond-control} and let $\Gm$ as in Def.~\ref{def:emp-op}, then
	\eqal{\label{eq:def-E}
		\nor{E} \leq \nor{\Gml^{-1/2}(\Cn - \Gm)\Gml^{-1/2}}.
	}
	\elm
	\bpr
	By multiplying and dividing  by $\Gml = \Gm + \la I$ we have
	\eqals{
		\nor{E} & =  \nor{A^{-\top} V^*\Gml^{1/2}~~\Gml^{-1/2}(\Cn - \Gm) \Gml^{-1/2}~~\Gml^{1/2} V A^{-1}} \\
		& \leq \nor{A^{-\top} V^*\Gml^{1/2}}^2 \nor{\Gml^{-1/2}(\Cn - \Gm) \Gml^{-1/2}}.
	}
	Now, considering that $V^*V = I$ and the identity in Eq.~\eqref{eq:AVCm-identity}, we have
	\eqal{\label{eq:proof-W-E-I}
		\nor{A^{-\top} V^*\Gml^{1/2}}^2 & = \nor{A^{-\top} V^*(\Gm + \la I) V A^{-1}} = \nor{A^{-\top} (V^*\Gm V + \la I)A^{-1}} = 1.
	}
	\epr
	
	\subsection{Analytic results (II): The computational oracle inequality}
	In this subsection (Lemma~\ref{lm:oracle-analytic}) we bound the excess risk of FALKON with respect to the one of the exact \Nystrom{} estimator. First we prove that FALKON is equal to the exact \Nystrom{} estimator as the iterations go to infinity (Lemma~\ref{lm:FALKON-in-H},~\ref{lm:Nystrom-in-H}). Then in Lemma~\ref{lm:oracle-analytic} (via Lemma~\ref{lm:oracle-help1},~\ref{lm:oracle-help2}) we use functional analytic tools, together with results from operator theory to relate the weak convergence result of the conjugate gradient method on the chosen preconditioned problem, with the excess risk.
	
	\blm[Representation of the FALKON estimator as vector in $\hh$]\label{lm:FALKON-in-H}
	Let $\la > 0$, $M, \iter \in \N$ and $B$ as in Def.~\ref{def:matr-algo}. The FALKON estimator as in Def.~\ref{def:generalized-FALKON} is characterized by the vector $\widehat{f} \in \hh$ as follows,
	\eqal{\label{eq:FALKON-general}
		\widehat{f}_{\la, M, \iter} ~~=~~ \sqrt{M}~ \Sm^* B \beta_{\iter},
	}
	where $\beta_\iter \in \R^q$ denotes the vector resulting from $\iter$ iterations of the conjugate gradient algorithm applied to the linear system in Def.~\ref{def:generalized-FALKON}.
	\elm
	\bpr
	According to the definition of $\widehat{f}_{\la, M, \iter}(\cdot)$ in Def.\ref{def:generalized-FALKON} and the definition of the operator $\Sm$ in Def.~\ref{def:emp-op}, denoting with $\alpha \in \R^M$ the vector $B \beta_{\iter}$, we have that
	$$\widehat{f}_{\la, M, \iter}(x) = \sum_{i=1}^M \alpha_i K(x,\tilde{x}_i) = \scal{K_x}{\sum_{i=1}^M \alpha_i K_{\tilde{x}_i}}_\hh = \scal{K_x}{~\sqrt{M}~\Sm^* \alpha}_\hh,$$
	for any $x \in \X$. Then the vector in $\hh$ representing the function $\widehat{f}_{\la, M, \iter}(\cdot)$ is
	$$ \widehat{f}_{\la, M, \iter} = \sqrt{M}~\Sm^* \alpha = \sqrt{M}~\Sm^* B \beta_t.$$
	\epr
	
	\blm[Representation of the \Nystrom{} estimator as a vector in $\hh$]\label{lm:Nystrom-in-H}
	Let $\la > 0, M \in \N$, and $B$ as in Def.~\ref{def:matr-algo}. The exact \Nystrom{} estimator, in Eq.\eqref{eq:form-subsampled} and Eq.~\eqref{eq:base-nystrom} is characterized by the vector $\widetilde{f} \in \hh$ as follows
	\eqal{\label{eq:FALKON-exact-nystrom}
		\widetilde{f}_{\la, M} ~~=~~ \sqrt{M}~\Sm^* B \beta_{\infty},
	}
	where $\beta_\infty = W^{-1} B^\top \Knm^\top \widehat{y}$ is the vector resulting from infinite iterations of the conjugate gradient algorithm applied to the linear system in Eq.~\eqref{eq:gen-FALKON-problem}.
	\elm
	\bpr
	For the same reasoning in the proof of Lemma~\ref{lm:FALKON-in-H}, we have that the FALKON estimator with infinite iterations is characterized by the following vector in $\hh$
	$$\widetilde{f}_{\la, M} = \sqrt{M}~\Sm^* B \beta_{\infty}.$$
	To complete the proof, we need to prove 1) that $\beta_\infty = W^{-1} B^\top \Knm \widehat{y}$ and 2) that $\widetilde{f}_{\la, M}$ above, corresponds to the exact \Nystrom{} estimator, as in Eq.~\eqref{eq:base-nystrom}.
	
	Now we characterize $\beta_\infty$. First, by the characterization of $W$ in Lemma~\ref{lm:characterization-W} and the fact that $V^*V = I$, we have
	\eqal{\label{eq:W-char}
		W = A^{-\top} V^*(\Cn + \la I) V A^{-1} = A^{-\top} (V^* \Cn V + \la I) A^{-1}.
	}
	Since $\Cn$ is a positive operator (see Def.~\ref{def:emp-op}) $A$ is invertible and $\la > 0$, then $W$ is a symmetric and positive definite matrix.
	The positive definiteness of $W$ implies that it is invertible and that is has a finite condition number, making the conjugate gradient algorithm to converge to the solution of the system in Eq.~\eqref{eq:gen-FALKON-problem} (Thm. 6.6 of \cite{saad2003iterative} and Eq.~6.107).
	So we can explicitly characterize $\beta_\infty$, by the solution of the system in Eq.~\eqref{eq:gen-FALKON-problem}, that is
	\eqal{\label{eq:beta-infty-charact}
		\beta_\infty = W^{-1} B^\top \Knm^\top \widehat{y}.
	}
	So we proved that $\widetilde{f}_{\la, M} \in \hh$, with the above characterization of $\beta_\infty$, corresponds to FALKON with infinite iterations. Now we show that $\widetilde{f}_{\la, M}$ is equal to the \Nystrom{} estimator given in \cite{rudi2015less}.
	First we need to study $\Sm^* B W^{-1} B^\top \Sm$. By the characterization of $W$ in Eq.~\eqref{eq:W-char},  the identity $(A B C)^{-1} = C^{-1}B^{-1}A^{-1}$, valid for any $A,B,C$ bounded invertible operators, and the definition of $V$ (Lemma~\ref{lm:characterization-W}),
	\eqal{\label{eq:Sm-BWB-Sm}
		\Sm^* B W^{-1} B^\top \Sm &= \Sm^* B \left(A^{-\top}(V^*\Cn V + \la I)A^{-1}\right)^{-1} B^\top \Sm \\
		& = \Sm^* B A (V^*\Cn V + \la I)^{-1} A^\top B^\top \Sm\\
		& = \frac{1}{Mn} V(V^*\Cn V + \la I)^{-1} V^*.
	}
	By expanding $\beta_\infty$, $\Knm$ (see Lemma~\ref{prop:basic_operator_result}) in $\widetilde{f}_{\la, M}$,
	\eqal{
		\widetilde{f}_{\la, M} &= \sqrt{M}~ \Sm^* B \beta_\infty = \sqrt{M} ~\Sm^* B W^{-1} B^\top \Knm^\top \yn =  \sqrt{n}M ~\Sm^* B W^{-1} B^\top \Sm\Sn^* \yn \\
		& = \frac{1}{\sqrt{n}}~ V (V^*\Cn V + \la I)^{-1} V^*\Sn^* \yn.
	}
	Now by Lemma 2 of \cite{rudi2015less} with $Z_m = \Sm$, we know that the exact \Nystrom{} solution is characterized by the vector $\bar{f} \in \hh$ defined as follows
	$$ \bar{f} = \frac{1}{\sqrt{n}} \bar{V}(\bar{V}^*\Cn \bar{V} + \la I)^{-1} \bar{V}^* \Sn^* \yn,$$
	with $\bar{V}$ a partial isometry, such that $\bar{V}^*\bar{V} = I$ and $\bar{V}\bar{V}^*$ with the same range of $\Sm^*$. Note that, by definition of $V$ in Lemma~\ref{lm:characterization-W}, we have that it is a partial isometry such that $V^*V = I$ and $VV^*$ with the same range of $\Sm^*$.
	This implies that $\bar{V} = VG$, for an orthogonal matrix $G \in \R^{q \times q}$.
	Finally, exploiting the fact that $G^{-1} = G^\top$, that $G G^\top = G^\top G = I$ and that for three invertible matrices $A, B, C$ we have $(A B C)^{-1} = C^{-1} B^{-1} A^{-1}$,
	\eqals{
		\bar{f} &= \frac{1}{\sqrt{n}} \bar{V}(\bar{V}^*\Cn \bar{V} + \la I)^{-1} \bar{V}^* \Sn^* \yn = \frac{1}{\sqrt{n}} VG\left(G^\top (V^*\Cn V + \la I)G\right)^{-1}  G^\top V^* \Sn^* \yn \\
		& = \frac{1}{\sqrt{n}} VG G^\top \left(V^*\Cn V + \la I\right)^{-1} G G^\top V^* \Sn^* \yn  = \frac{1}{\sqrt{n}} V \left(V^*\Cn V + \la I\right)^{-1} V^* \Sn^* \yn = \widetilde{f}_{\la, M}.
	}
	\epr
	
	The next lemma is necessary to prove Lemma~\ref{lm:oracle-analytic}.
	\blm\label{lm:oracle-help1}
	When $\la > 0$ and $B$ is as in Def.~\ref{def:matr-algo}.
	then
	$$\sqrt{M}\nor{S \Sm^* B W^{-1/2}} \leq n^{-1/2} \nor{S \Cnl^{-1/2}}.$$
	\elm
	\bpr
	By the fact that
	identity $\nor{Z}^2 = \nor{ZZ^*}$ valid for any bounded operator $Z$ and the identity in Eq.~\ref{eq:Sm-BWB-Sm}, we have
	\eqals{
		M\nor{S \Sm^* B W^{-1/2}}^2 & = M\nor{S \Sm^* B W^{-1} B^\top \Sm S^*} = \frac{1}{n} \nor{S V (V^*\Cn V + \la I)^{-1} V^* S^*} \\
		& = \frac{1}{n}\nor{S V (V^*\Cn V + \la I)^{-1/2}}^2.
	}
	Denote with $\Cnl$ the operator $\Cn + \la I$, by dividing and multiplying for $\Cnl^{-1/2}$, we have
	$$ S V (V^*\Cn V + \la I)^{-1/2} = S\Cnl^{-1/2} ~~ \Cnl^{1/2} V (V^*\Cn V + \la I)^{-1/2}.$$
	The second term is equal to $1$, indeed, since $V^*\Cnl V = V^*\Cn V + \la I$, and $\nor{Z}^2 = \nor{Z^*Z}$, for any bounded operator $Z$, we have
	\eqal{\label{eq:VCnVSn}
		\nor{\Cnl^{1/2} V (V^*\Cn V + \la I)^{-1/2}}^2 &= \nor{(V^*\Cn V + \la I)^{-1/2} V^*\Cnl V (V^*\Cn V + \la I)^{-1/2}}\\
		& = \nor{(V^*\Cn V + \la I)^{-1/2}(V^*\Cn V + \la I) (V^*\Cn V + \la I)^{-1/2}} \\
		& = 1.
	}
	Finally
	\eqals{
		\sqrt{M}\nor{S \Sm^* B W^{-1/2}} & = \frac{1}{\sqrt{n}}\nor{S V (V^*\Cn V + \la I)^{-1/2}} \\
		& \leq \frac{1}{\sqrt{n}}\nor{S \Cnl^{-1/2}} \nor{\Cnl^{1/2} V (V^*\Cn V + \la I)^{-1/2}} \\
		& \leq n^{-1/2} \nor{S \Cnl^{-1/2}}.
	}
	\epr

	The next lemma is necessary to prove Lemma~\ref{lm:oracle-analytic}.
	\blm\label{lm:oracle-help2}
	For any $\la > 0$, let $\beta_\infty$ be the vector resulting from infinite iterations of the conjugate gradient algorithm applied to the linear system in Eq.~\eqref{eq:gen-FALKON-problem}. Then
	$$\nor{W^{1/2}\beta_\infty}_{\R^q} \leq \nor{\yn}_{\R^n}.$$
	\elm
	\bpr
	First we recall the characterization of $\beta_\infty$ from Lemma~\ref{lm:Nystrom-in-H},
	$$ \beta_\infty = W^{-1} B^\top \Knm^\top \yn.$$
	So, by the characterization of $\Knm$ in terms of $\Sn, \Sm$ (Prop.~\ref{prop:basic_operator_result}),
	\eqals{
		W^{1/2} \beta_\infty = W^{1/2} W^{-1} B^\top \Knm^\top \yn = \sqrt{nM}~ W^{-1/2} B^\top \Sm \Sn^* \yn.
	}
	Then, by applying the characterization of $\Sm^* B W^{-1} B^\top \Sm$ in terms of $V$, in Eq.~\ref{eq:Sm-BWB-Sm}
	\eqals{
		\nor{W^{1/2} \beta_\infty}^2_{\R^q} & = nM ~~ \nor{W^{-1/2} B^\top \Sm \Sn^*\yn}^2_{\R^q} = nM ~ \yn^\top \Sn \Sm^* B W^{-1} B^\top \Sm \Sn^* \yn \\
		& = \yn^\top \Sn V(V^*\Cn V + \la I)^{-1} V^* \Sn^* \yn = \nor{(V^*\Cn V + \la I)^{-1/2}  V^* \Sn^* \yn}^2_{\R^q}.
	}
	Finally
	$$ \nor{(V^*\Cn V + \la I)^{-1/2} \Sn^* \yn}_{\R^q} \leq \nor{(V^*\Cn V + \la I)^{-1/2} \Sn^*} \nor{\yn}_{\R^n}. $$
	Note that
	$$ \nor{(V^*\Cn V + \la I)^{-1/2} \Sn^*} \leq 1,$$
	indeed
	$$ \nor{(V^*\Cn V + \la I)^{-1/2} \Sn^*} \leq \nor{(V^*\Cn V + \la I)^{-1/2} \Cnl^{1/2}}\nor{\Cnl^{-1/2} \Sn^*},$$
	and the first term is equal to 1 by Eq.~\eqref{eq:VCnVSn}, moreover by definition of $\Cn$ (Def.~\ref{def:emp-op}),
	$$ \nor{\Cnl^{-1/2} \Sn^*}^2 = \nor{\Cnl^{-1/2} \Cn \Cnl^{-1/2}} = \nor{\Cnl^{-1/2}\Cn^{1/2}}^2 = \sup_{\sigma \in \sigma(\Cn)} \frac{\sigma}{\sigma + \la} \leq 1,$$
	where $\sigma(\Cn) \subset [0, \nor{\Cn}]$ is the set of eigenvalues of $\Cn$.
	\epr
	
	\blm\label{lm:oracle-analytic}
	Let $M \in \N$, $\la > 0$ and $B$ satisfying Def.~\ref{def:generalized-FALKON}. Let $\widehat{f}_{\la, M, \iter}$ be the FALKON estimator after $\iter \in \N$ iterations and $\widetilde{f}_{\la, M}$ the exact \Nystrom{} estimator as in Eq.~\ref{eq:form-subsampled},~\ref{eq:base-nystrom}. Let $c_0 \geq 0$ such that
	$$\nor{S \Cnl^{-1/2}} \leq c_0,$$
	then
	$$
	{\cal R}(\widehat{f}_{\la, M, \iter})^{1/2} ~~\leq~~ {\cal R}(\widetilde{f}_{\la, M}~)^{1/2} ~~ + ~~ 2 c_0 ~\widehat{v}~ \left(1 - \frac{2}{\sqrt{\textrm{cond(W)}} + 1}\right)^{\iter},
	$$
	where $\widehat{v}^2 = \frac{1}{n} \sum_{i=1}^n y_i^2$.
	\elm
	\bpr[Proof of Lemma~\ref{lm:oracle-analytic}]
	By Prop.~\ref{prop:gen-error} we have that for any $f \in \hh$
	$$({\cal E}(f) - \inf_{f \in \hh} {\cal E}(f))^{1/2} = \nor{Sf - P\frho}_\rhox,$$
	with $P: \Ltwo \to \Ltwo$ the orthogonal projection operator whose range is the closure of the range of $S$ in $\Ltwo$.
	Let $\widehat{f}_{\la, M,\iter} \in \hh$ and $\widetilde{f}_{\la, M} \in \hh$ be respectively the Hilbert vector representation of the FALKON estimator and of the exact \Nystrom{} estimator (Lemma~\ref{lm:FALKON-in-H} and Lemma~\ref{lm:Nystrom-in-H}). By adding and subtracting $\widetilde{f}_{\la, M}$ we have
	\eqals{
		|{\cal E}(\widehat{f}~) - \inf_{f \in \hh} {\cal E}(f)|^{1/2} &= \nor{S\widehat{f}_{\la, M, \iter} -  P\frho}_\rhox = \nor{S(\widehat{f}_{\la, M, \iter} - \widetilde{f}_{\la, M}) ~+~ (S\widetilde{f}_{\la, M} - P\frho)}_\rhox\\
		& \leq \nor{S(\widehat{f}_{\la, M, \iter} - \widetilde{f}_{\la, M})}_\rhox + \nor{S\widetilde{f}_{\la, M} - P\frho}_\rhox \\
		& = \nor{S(\widehat{f}_{\la, M, \iter} - \widetilde{f}_{\la, M})}_\rhox + |{\cal E}(\widetilde{f}_{\la, M}) - \inf_{f \in \hh} {\cal E}(f)|^{1/2}.
	}
	In particular, by expanding the definition of $\widehat{f}_{\la, M, \iter}, \widetilde{f}_{\la, M}$ from  Lemma~\ref{lm:FALKON-in-H} and Lemma~\ref{lm:Nystrom-in-H}, we have
	$$\nor{S(\widehat{f}_{\la, M, \iter} - \widetilde{f}_{\la, M})}_\rhox = \sqrt{M}\nor{S \Sm^* B(\beta_\iter - \beta_\infty)}_\rhox,$$
	where $\beta_\iter \in \R^q$ and $\beta_\infty \in \R^q$ denote respectively the vector resulting from $\iter$ iterations and infinite iterations of the conjugate gradient algorithm applied to the linear system in Eq.~\eqref{eq:gen-FALKON-problem}.
	Since $W$ is symmetric positive definite when $\la > 0$ (see proof of Lemma~\ref{lm:Nystrom-in-H}), we can apply the standard convergence results for the conjugate gradient algorithm (Thm. 6.6 of \cite{saad2003iterative}, in particular Eq.~6.107), that is the following
	$$\nor{W^{1/2}(\beta_\iter - \beta_\infty)}_{\R^q} \leq q(W,\iter) \nor{W^{1/2}\beta_\infty}_{\R^q},\quad \textrm{with} \quad q(W,\iter) = 2\left(1 - \frac{2}{\sqrt{\textrm{cond(W)}} + 1}\right)^{\iter}.$$
	So by dividing and multiplying by $W^{1/2}$ we have
	\eqals{
		\nor{S(\widehat{f}_{\la, M, \iter} - \widetilde{f}_{\la, M})}_\rhox &= \sqrt{M}\nor{S \Sm^* B(\beta_\iter - \beta_\infty)}_\rhox = \sqrt{M}\nor{S \Sm^* BW^{-1/2}W^{1/2}(\beta_\iter - \beta_\infty)}_\rhox\\
		& \leq \sqrt{M}\nor{S \Sm^* B W^{-1/2}} \nor{W^{1/2}(\beta_\iter - \beta_\infty)}_{\R^q} \\
		& \leq q(W, \iter)~ \sqrt{M}\nor{S \Sm^* B W^{-1/2}} \nor{W^{1/2}\beta_\infty}_{\R^q}.
	}
	Finally, the term $\sqrt{M}\nor{S \Sm^* B W^{-1/2}}$ is bounded in Lemma~\ref{lm:oracle-help1} as
	$$\sqrt{M}\nor{S \Sm^* B W^{-1/2}} \leq \frac{1}{\sqrt{n}}\nor{S \Cnl^{-1/2}} \leq \frac{c_0}{\sqrt{n}},$$
	while, for the term $\nor{W^{1/2}\beta_\infty}_{\R^q}$, by Lemma~\ref{lm:oracle-help2}, we have
	$$\nor{W^{1/2}\beta_\infty}_{\R^q} \leq \nor{\yn}_{\R^{n}}  =  (\sum_{i=1} y_i^2)^{1/2} = \sqrt{n} \sqrt{\frac{\sum_{i=1}^n y_i^2}{n}}  = \sqrt{n} \widehat{v}.$$
	\epr
	
	\section{Probabilistic Estimates}\label{sect:prob-est}
	
	In Lemma~\ref{lm:ny-uni},~\ref{lm:ny-als} we provide probabilistic estimates of $\nor{E}$, the quantity needed to bound the condition number of the preconditioned linear system of FALKON (see Lemma~\ref{lm:characterization-W},~\ref{lm:control-E}). In particular Lemma~\ref{lm:ny-uni}, analyzes the case when the \Nystrom{} centers are selected with uniform sampling, while Lemma~\ref{lm:ny-als}, considers the case when the \Nystrom{} centers are selected via approximate leverage scores sampling.

	Now we are ready to provide probabilistic estimates for uniform sampling.
	
	\blm\label{lm:ny-uni}
	Let $\eta \in [0,1)$ and $\delta \in (0, 1]$. When $\tilde{x}_1, \dots, \tilde{x}_M$ are selected via \Nystrom{} uniform sampling (see Sect.~\ref{sect:gen-algo}), $0 < \la \leq \nor{C}$, $M \leq n$ and
	\eqal{\label{eq:uni-nyst-beta2}
		M \geq 4\left[\frac{1}{2} + \frac{1}{\eta} + \left(\frac{3 + 7 \eta}{3 + 3 \eta}\right)\left(1 + \frac{2}{\eta}\right)^2{\cal N}_\infty(\la)\right]\log \frac{8\kappa^2}{\la \delta},
	}
	then the following hold with probability at least $1 - \delta$,
	$$\nor{\Cl^{-1/2}(C - \Cn)\Cl^{-1/2}} < \eta, \quad \nor{\Gml^{-1/2}(\Cn - \Gm)\Gml^{-1/2}} < \eta.$$
	\elm
	\bpr
	First of all, note that since the \Nystrom{} centers are selected by uniform sampling. Then $\tilde{x}_1, \dots, \tilde{x}_M$ are independently and identically distributed according to $\rhox$ and moreover $D$ is the identity matrix. So 
	$$\Gm = \Sm^*D^2 \Sm = \Sm^*\Sm = \Cm.$$
	Note that, by multiplying and dividing by $\Cl$,
	\eqals{
		\nor{\Gml^{-1/2}(\Cn - \Gm)\Gml^{-1/2}} & = \nor{\Cml^{-1/2}(\Cn - \Cm)\Cml^{-1/2}} \\ 
		& = \nor{\Cml^{-1/2}\Cl^{1/2}\Cl^{-1/2}(\Cn - \Cm)\Cl^{-1/2}\Cl^{1/2}\Cml^{-1/2}} \\
		& \leq \nor{\Cml^{-1/2}\Cl^{1/2}}^2\nor{\Cl^{-1/2}(\Cn - \Cm)\Cl^{-1/2}} \\
		& \leq (1 - \la_{\max}(\Cl^{-1/2}(C - \Cm)\Cl^{-1/2}))^{-1}\nor{\Cl^{-1/2}(\Cn - \Cm)\Cl^{-1/2}}
	}
	where the last step is due to Prop. 9 of \cite{rudi2016rf}. Moreover note that
	$$\la_{\max}(\Cl^{-1/2}(C - \Cm)\Cl^{-1/2}) \leq \nor{\Cl^{-1/2}(C - \Cm)\Cl^{-1/2}}.$$
	
	Let $\mu = \frac{\delta}{2}$. Note that $\Cm = \frac{1}{M}\sum_{i=1}^M v_i \otimes v_i$ with $v_i$ the random variable $v_i = K_{\tilde{x}_i}$ (see Prop.~\ref{prop:basic_operator_result}) and, since $\tilde{x}_1, \dots, \tilde{x}_M$ are i.i.d. w.r.t. $\rhox$, by the characterization of $C$ in  Prop.~\ref{prop:basic_operator_result}, we have for any $1 \leq i \leq M$,
	$$\mathbb{E} v_i \otimes v_i = \int_{\X} K_{x} \otimes K_x d\rhox(x) = C.$$
	Then, by considering that $\nor{v} = \nor{K_x} \leq \kappa^2$ a. e., we can apply Prop.~7 of \cite{rudi2016rf}, obtaining
	$$ \nor{\Cl^{-1/2}(C - \Cm)\Cl^{-1/2}} \leq \frac{2d(1 + {\cal N}_\infty(\la))}{3M} + \sqrt{\frac{2d{\cal N}_\infty(\la)}{3M}}, \quad d = \log\frac{4\kappa^2}{\la \mu},$$
	with probability at least $1 - \mu$.
	Note that, when $M$ satisfies Eq~\eqref{eq:uni-nyst-beta2}, we have
	$ \nor{\Cl^{-1/2}(C - \Cm)\Cl^{-1/2}} < \eta/(2 + \eta)$.
	By repeating the same reasoning for $C_n$, we have
	$$ \nor{\Cl^{-1/2}(C - \Cn)\Cl^{-1/2}} \leq \frac{2d(1 + {\cal N}_\infty(\la))}{3n} + \sqrt{\frac{2d{\cal N}_\infty(\la)}{3n}}, \quad d = \log\frac{4\kappa^2}{\la \mu},$$
	with probability $1 - \mu$.
	Since $n \geq M$ and $M$ satisfying Eq.~\eqref{eq:uni-nyst-beta2}, we have automatically that $\nor{\Cl^{-1/2}(C - \Cn)\Cl^{-1/2}} < \eta/(2 + \eta)$.
	
	Finally note that, by adding and subtracting $C$,
	\eqals{
		\nor{\Cl^{-1/2}(\Cn - \Cm)\Cl^{-1/2}} & = \nor{\Cl^{-1/2}((\Cn - C) + (C - \Cm))\Cl^{-1/2}}\\
		&  \leq \nor{\Cl^{-1/2}(C - \Cn)\Cl^{-1/2}} + \nor{\Cl^{-1/2}(C - \Cm)\Cl^{-1/2}}.
	}
	So by performing the intersection bound of the two previous events, we have
	\eqals{
		\|\Cml^{-1/2}&(\Cn - \Cm)\Cml^{-1/2}\| \leq (1 - \nor{\Cl^{-1/2}(C - \Cn)\Cl^{-1/2}})^{-1} \times\\ & \times \left(\nor{\Cl^{-1/2}(C - \Cn)\Cl^{-1}} + \nor{\Cl^{-1/2}(C - \Cm)\Cl^{-1/2}}\right) < \eta,
	}
	with probability at least $1 - 2\mu$. The last step consists in substituting $\mu$ with $\delta/2$.
	\epr
	
	The next lemma gives probabilistic estimates for $\nor{E}$, that is the quantity needed to bound the condition number of the preconditioned linear system of FALKON (see Lemma~\ref{lm:characterization-W},~\ref{lm:control-E}), when the \Nystrom{} centers are selected via approximate leverage scores sampling.
	
	\blm\label{lm:ny-als}
	Let $\eta > 0$, $\delta \in (0, 1]$, $n, M \in \N$, $q \geq 1$ and $\la_0 > 0$. Let $x_1,\dots,x_n$ be independently and identically distributed according to $\rhox$. 
	Let $\tilde{x}_1, \dots, \tilde{x}_M$ be randomly selected from $x_1 \dots, x_n$, by using the $(q,\la_0, \delta)$-approximate leverage scores (see Def.~\ref{def:emp-q-als} and discussion below), with $\la_0 \vee \frac{19\kappa^2}{n} \log \frac{n}{2\delta} \leq \la \leq \nor{C}$. When $n \geq 405 \kappa^2 \vee 67 \kappa^2 \log \frac{12\kappa^2}{\delta}$ and 
	\eqal{\label{eq:lev-nyst-beta}
		M \geq \left[2 + \frac{2}{\eta} + \frac{18(\eta^2 + 5\eta + 4) q^2}{\eta^2}{\cal N}(\la) \right]\log \frac{8\kappa^2}{\la \delta},
	}
	then the following hold with probability at least $1 - \delta$,
	$$\nor{\Gml^{-1/2}(\Cn - \Gm)\Gml^{-1/2}} < \eta, \quad \nor{\Cl^{-1/2}(C - \Cn)\Cl^{-1/2}} < \eta.$$
	\elm
	\bpr
	By multiplying and dividing by $\Cnl = \Cn + \la I$, we have
	\eqals{
		\nor{\Gml^{-1/2}(\Cn - \Gm)\Gml^{-1/2}} & = \nor{\Gml^{-1/2}\Cnl^{1/2}\Cnl^{-1/2}(\Cn - \Gm)\Cnl^{-1/2}\Cnl^{1/2}\Gml^{-1/2}} \\
		& \leq \nor{\Gml^{-1/2}\Cnl^{1/2}}^2\nor{\Cnl^{-1/2}(\Cn - \Gm)\Cnl^{-1/2}} \\
		& \leq (1 - \la_{\max}(\Cnl^{-1/2}(\Cn - \Gm)\Cnl^{-1/2}))^{-1}\nor{\Cnl^{-1/2}(\Cn - \Gm)\Cnl^{-1/2}}
	}
	where the last step is due to Prop.~9 of \cite{rudi2016rf}. Note that
	$$\la_{\max}(\Cnl^{-1/2}(\Cn - \Gm)\Cnl^{-1/2}) \leq \nor{\Cnl^{-1/2}(\Cn - \Gm)\Cnl^{-1/2}},$$
	thus
	$$
	\nor{\Gml^{-1/2}(\Cn - \Gm)\Gml^{-1/2}} \leq \frac{t}{1-t},
	$$
	with $t = \nor{\Cnl^{-1/2}(\Cn - \Gm)\Cnl^{-1/2}}$. Now we bound $t$.
	We denote with $l_\la(j)$, $\widehat{l}_\la(j)$, respectively the leverage scores and the $(q,\la_0,\delta)$-approximate leverage score associated to the point $x_j$, as in Def.~\ref{def:emp-q-als} and discussion above.
	First we need some considerations on the leverage scores. By the spectral theorem and the fact that $K_{nn} = n~\Sn\Sn^*$ (see Prop.~\ref{prop:basic_operator_result}), we have
	\eqals{
		l_\la(j) &= (\K (\K + \la n I)^{-1})_{jj} = e_j^\top \Sn\Sn^*(\Sn\Sn^* + \la I)^{-1} e_j = e_j^\top \Sn(\Sn^*\Sn + \la I)^{-1}\Sn^* e_j \\
		& = \frac{1}{n}\scal{K_{x_j}}{\Cnl^{-1}K_{x_j}} = \frac{1}{n}\nor{\Cnl^{-1/2}K_{x_j}}^2.
	}
	for any $1 \leq j \leq n$. Moreover, by the characterization of $\Cn$ in Prop.~\ref{prop:basic_operator_result}, we have
	\eqals{
		\frac{1}{n}\sum_{j=1}^n l_\la(j) &= \frac{1}{n}\sum_{j=1}^n \scal{K_{x_j}}{(\Cn+\la)^{-1} K_{x_j}}_\hh = \frac{1}{n}\sum_{j=1}^n \tr((\Cn+\la)^{-1} (K_{x_j} \otimes K_{x_j})) \\
		& = \tr((\Cn+\la)^{-1} \frac{1}{n}\sum_{j=1}^n (K_{x_j} \otimes K_{x_j})) = \tr(\Cnl^{-1}\Cn).
	}
	Since the \Nystrom{} points are selected by using the $(q,\la_0,\delta)$-approximate leverage scores, then $\widetilde{x}_t = x_{i_t}$ for $1 \leq t \leq M$, where $i_1,\dots, i_M \in \{1,\dots,n\}$ is the sequence of indexes obtained by approximate leverage scores sampling (see Sect.~\ref{sect:gen-algo}). Note that $i_1,\dots, i_M$ are independent random indexes, distributed as follows: for $1 \leq t \leq M$,
	$$i_t = j, \quad \textrm{with probability} \quad p_j = \frac{\widehat{l}_\la(j)}{\sum_{h=1}^n \widehat{l}_\la(h)}, \quad \forall ~ 1 \leq j \leq n.$$
	Then, by recalling the definition of $\Gm$ with respect to the matrix $D$ defined as in Def.~\ref{def:diag-D} and by Prop.~\ref{prop:basic_operator_result} we have,
	$$\Gm = \Sm^*D^2\Sm =  \frac{1}{M } \sum_{t=1}^M \frac{1}{n p_{i_t}} K_{x_{i_t}} \otimes K_{x_{i_t}}.$$
	Consequently $\Gm = \frac{1}{M} \sum_{i=1}^M v_i \otimes v_i$, where $(v_i)_{i=1}^M$ are independent random variables distributed in the following way
	$$v_i = \frac{1}{\sqrt{p_j n}} K_{x_j}, \quad \textrm{with probability} ~~  p_j, \quad \forall~ 1 \leq j \leq n.$$
	
	Now we study the moments of $\Gm$ as a sum of independent random matrices, to apply non-commutative Bernstein inequality (e.g. Prop.~7 of \cite{rudi2016rf}). We have that, for any $1 \leq i \leq M$
	\eqals{
		{\mathbb E} v_i \otimes v_i& = \sum_{j=1}^n p_j \left(\frac{1}{p_j n} K_{x_j} \otimes K_{x_j}\right) = \Cn,\\
		\scal{v_i}{\Cnl^{-1}v_i}_\hh & \leq \sup_{1 \leq j \leq n} \frac{\nor{\Cnl^{-1/2} K_{x_j}}^2}{p_j n} = \sup_{1 \leq j \leq n} \frac{l_\la(j)}{p_j n}
		= \sup_{1 \leq j \leq n} \frac{l_\la(j)}{\widehat{l}_\la(j)} \frac{1}{n} \sum_{h=1}^n \widehat{l}_\la(h) \\
		& \leq q \frac{1}{n}\sum_{h=1}^n \widehat{l}_\la(h) \leq q^2 \frac{1}{n}\sum_{h=1}^n l_\la(h) = q^2 \tr (\Cnl^{-1} \Cn),
	}
	for all $1 \leq j \leq n$.
	Denote with $\widehat{\cal N}(\la)$, the quantity $\tr(\Cnl^{-1}\Cn)$, by applying Prop.~7 of \cite{rudi2016rf}, we have
	\eqals{
		\nor{\Cnl^{-1/2}(\Cn - \Gm)\Cnl^{-1/2}} \leq \frac{2 d(1 + q^2\widehat{\cal N}(\la))}{3M} + \sqrt{\frac{2 d q^2\widehat{\cal N}(\la)}{M}}, \quad d = \log \frac{\kappa^2}{\la \mu}.
	}
	with probability at least $1 - \mu$. The final step consist in bounding the empirical intrinsic dimension $\widehat{\cal N}(\la)$ with respect to intrinsic dimension ${\cal N}(\la)$, for which we use Prop.~1 of \cite{rudi2015less}, obtaining
	$$ \widehat{\cal N}(\la) \leq 2.65{\cal N}(\la),$$
	with probability at least $1 - \mu$, when $n \geq 405 \kappa^2 \vee 67 \kappa^2 \log \frac{6\kappa^2}{\mu}$ and $\frac{19\kappa^2}{n} \log \frac{n}{4\mu} \leq \la \leq \nor{C}$.
	By intersecting the events, we have
	\eqals{
		\nor{\Cnl^{-1/2}(\Cn - \Gm)\Cnl^{-1/2}} \leq \frac{5.3 d(1 + q^2{\cal N}(\la))}{3M} + \sqrt{\frac{5.3 d q^2{\cal N}(\la)}{M}}, \quad d = \log \frac{\kappa^2}{\la \mu}.
	}
	with probability at least $1 - 2\mu$. The last step consist in substituting $\mu$ with $\mu = \delta/2$. Thus, by selecting $M$ as in Eq.~\ref{eq:lev-nyst-beta}, we have
	\eqals{
		t = \nor{\Cnl^{-1/2}(\Cn - \Gm)\Cnl^{-1/2}} < \frac{\eta}{1 + \eta}.
	}
	That implies, 
	\eqals{
		\nor{\Gml^{-1/2}(\Cn - \Gm)\Gml^{-1/2}} < \frac{t}{1-t} < \eta.
	}
	\epr
	
	\section{Proof of Main Results}\label{sect:proof-main-res}
	
	In this section we prove the main results of the paper. This section is divided in three subsections. In the first, we specify the computational oracle inequality for \Nystrom{} with uniform sampling, in the second we specify the computational oracle inequality for \Nystrom{} with approximate leverage scores sampling (see Sect.~\ref{sect:prob-est} for a definition), while the third subsection contains  the proof of the main theorem presented in the paper.
	
	Now we give a short sketch of the structure of the proofs. The definition of the general version of the FALKON algorithm (taking into account leverage scores and non invertible $\Km$) is given in Sect.~\ref{sect:gen-algo}. In Sect.~\ref{sect:notation} the notation and basic definition required for the rest of the analisys are provided.
	
	Our starting point is the analysis of the basic \Nystrom{} estimator given in \cite{rudi2015less}.
	The key novelty is the  quantification of the approximations induced by the preconditioned iterative solver by relating its  excess risk to the one of the basic \Nystrom{} estimator.
	
	{\em A computational oracle inequality.}  First we prove that FALKON is equal to the exact \Nystrom{} estimator as the iterations go to infinity (Lemma~\ref{lm:Nystrom-in-H}, Sect.~\ref{sect:analytic-dec}). \loz{Then, in Lemma~\ref{lm:oracle-analytic} (see also  Lemma~\ref{lm:oracle-help1},~\ref{lm:oracle-help2}, Sect.~\ref{sect:analytic-dec}) we show how optimization guarantees can be used to derive statistical results. More precisely, while optimization results  in machine learning typically derives guarantees on empirical minimization problems, we show, using analytic and probabilistic tools,  how these results can be turned into guarantees on the expected risks. }
	Finally, in the proof of Thm.~\ref{thm:oracle-inequality-base} we concentrate the terms of the inequality. The other key point is the study of the behavior of the condition number of $B^\top H B$ with $B$ given in~\eqref{eq:B-base}.
	
	{\em Controlling the condition number of $B^\top H B$}. Let $C_n$, $C_M$ be the empirical correlation operators in $\hh$ associated respectively to the training set and the \Nystrom{} points
	$C_n =\frac{1}{n} \sum_{i=1}^n K_{x_i} \otimes K_{x_i}$, $C_M = \frac{1}{M} \sum_{j=1}^M K_{\widetilde{x}_j} \otimes K_{\widetilde{x}_j}.$
	In Lemma~\ref{lm:characterization-W}, Sect.~\ref{sect:analytic-dec}, we prove that $B^\top H B$ is equivalent to $A^{-\top} V^* (C_n + \la I) V A^{-1}$ for a suitable partial isometry $V$. Then in Lemma~\ref{lm:cond-control}, Sect.~\ref{sect:analytic-dec}, we split it in two components
	\eqal{\label{eq:sketch-proof-split-BHB}
		B^\top H B = A^{-\top} V^* (C_M + \la I) V A^{-1} ~+~  A^{-\top} V^* (C_n - C_M) V A^{-1},
	}
	and prove that the first component is just the identity matrix. By denoting the second component with $E$, Eq.~\eqref{eq:sketch-proof-split-BHB}, Sect.~\ref{sect:analytic-dec}, implies that the condition number of $B^\top H B$ is bounded by $(1 + \nor{E})/(1 - \nor{E})$, when $\nor{E} < 1$. In Lemma~\ref{lm:control-E} we prove that $\nor{E}$ is analytically bounded by a suitable distance between $C_n - C_M$ and in Lemma~\ref{lm:ny-uni},~\ref{lm:ny-als}, Sect.~\ref{sect:prob-est}, we bound in probability such distance, when the \Nystrom{} centers are selected uniformly at random and with approximate leverage scores. Finally in Lemma~\ref{lm:ny-uni-controls-W},~\ref{lm:ny-als-controls-W}, Sect.~\ref{sect:prob-est}, we give a condition on $M$ for the two kind of sampling, such that the condition number is controlled and the error term in the oracle inequality decays as $e^{-t/2}$, leading to Thm.~\ref{thm:M-simple-rates},~\ref{thm:M-fast-rates}.
	
	Now we provide the preliminary result necessary to prove a computational oracle inequality for FALKON.
	
	{\bf Theorem~\ref{thm:oracle-inequality-base}}~~{\em
		Let $0 \leq \la \leq \nor{C}$, $B$ as in Def.~\ref{def:matr-algo} and $n, M, t\in \N$. Let $\widehat{f}_{\la, M, \iter}$ be the FALKON estimator, with preconditioner $B$, after $\iter$ iterations Def.~\ref{def:generalized-FALKON} and let $\widetilde{f}_{\la, M}$ be the exact \Nystrom{} estimator as in Eq.~\ref{eq:base-nystrom}. Let $\delta \in (0, 1]$ and $n \geq 3$, then following holds with probability $1 - \delta$
		$$
		{\cal R}(\widehat{f}_{\la, M, \iter})^{1/2} ~~\leq~~ {\cal R}(\widetilde{f}_{\la, M}~)^{1/2} ~~ + ~~ 4\widehat{v} ~ e^{- \nu \iter} ~ \sqrt{1 + \frac{9\kappa^2}{\la n} \log \frac{n}{\delta}},
		$$
		where $\widehat{v}^2 = \frac{1}{n} \sum_{i=1}^n y_i^2$ and $\nu = \log \frac{\sqrt{\cond{W}} + 1}{\sqrt{\cond{W}} - 1}$. In particular $\nu \geq 1/2$, when $\cond{W} \leq (\frac{e^{1/2} + 1}{e^{1/2} - 1})^2$.
	}
	\bpr
	By applying Lemma~\ref{lm:oracle-analytic}, we have
	$$
	{\cal R}(\widehat{f}_{\la, M, \iter})^{1/2} ~~\leq~~ {\cal R}(\widetilde{f}_{\la, M}~)^{1/2} ~~ + ~~ 2 c_0\nor{S \Cnl^{-1/2}} ~\widehat{v}~ e^{-\nu \iter}.
	$$
	To complete the theorem we need to study the quantity $\nor{S\Cnl^{-1/2}}$. In particular, define $\la_0 = \frac{9\kappa^2}{n}\log\frac{n}{\delta}$. By dividing and multiplying for $C_{n \la_0}^{1/2}$, we have
	$$\nor{S\Cnl^{-1/2}} = \nor{S C_{n \la_0}^{-1/2} C_{n \la_0}^{1/2} C_{n \la}^{-1/2}} \leq \nor{S C_{n \la_0}^{-1/2}} \nor{C_{n \la_0}^{1/2} C_{n \la}^{-1/2}}.$$
	Now, for the first term, since $\nor{Z}^2 = \nor{Z^*Z}$, and the fact that $C = S^*S$ (see Prop.~\ref{prop:basic_operator_result}), we have
	$$ \nor{S C_{n \la_0}^{-1/2}}^2 = \nor{C_{n \la_0}^{-1/2} C C_{n \la_0}^{-1/2}} = \nor{C^{1/2}C_{n \la_0}^{-1/2}},$$
	moreover by Lemma~5 of \cite{rudi2015less} (or Lemma~7.6 of \cite{rudi2013sample}), we have
	$$\nor{C^{1/2}C_{n \la_0}^{-1/2}} \leq 2,$$
	with probability $1-\delta$.
	Finally, by denoting with $\sigma(C)$ the set of eigenvalues of the positive operator $C$, recalling that $\sigma(C) \subset [0, \kappa^2]$ (see Prop.~\ref{prop:basic_operator_result}), we have
	$$\nor{C_{n \la_0}^{1/2} C_{n \la}^{-1/2}} = \sup_{\sigma \in \sigma(C)} \sqrt{\frac{\sigma + \la_0}{\sigma+\la}} \leq \sup_{\sigma \in [0, \kappa^2]} \sqrt{\frac{\sigma + \la_0}{\sigma+\la}} \leq \sqrt{1 + \frac{\la_0}{\la}}.$$
	\epr
	
	\subsection{Main Result (I): computational oracle inequality for FALKON with uniform sampling}
	
	\blm\label{lm:ny-uni-controls-W}
	Let $\delta \in (0,1]$, $0 < \la \leq \nor{C}$, $n, M \in \N$, the matrix $W$ as in Eq.~\ref{eq:gen-FALKON-problem} with $B$ satisfying Def.~\ref{def:matr-algo} and the \Nystrom{} centers selected via uniform sampling.
	When
	\eqal{\label{eq:ny-uni-cond-M-final}
		M \geq 5\left[1 + 14{\cal N}_\infty(\la)\right]\log \frac{8\kappa^2}{\la \delta},
	}
	then the following holds with probability $1-\delta$
	$$
	\cond{W} \leq \left(\frac{e^{1/2} + 1}{e^{1/2} - 1}\right)^2.
	$$
	\elm
	\bpr
	By Lemma~\ref{lm:characterization-W} we have that
	$$\cond{W} \leq \frac{1 + \nor{E}}{1 - \nor{E}},$$
	with the operator $E$ defined in the same lemma. By Lemma~\ref{lm:control-E}, we have
	$$\nor{E} \leq \nor{\Gml^{-1/2}(\Cn - \Gm)\Gml^{-1/2}}.$$
	Lemma~\ref{lm:ny-uni} proves that when the \Nystrom{} centers are selected with uniform sampling and $M$ satisfies Eq.~\eqref{eq:uni-nyst-beta2} for a given parameter $\eta \in (0,1]$, then $\nor{\Gml^{-1/2}(\Cn - \Gm)\Gml^{-1/2}} \leq \eta$, with probability $1 - \delta$. In particular we select $\eta = \frac{2e^{1/2}}{e + 1}$. The condition on $M$ in Eq.~\eqref{eq:ny-uni-cond-M-final} is derived by Eq.~\eqref{eq:uni-nyst-beta2} by substituting $\eta$ with $\frac{2e^{1/2}}{e + 1}$.
	\epr

	\bt\label{thm:oracle-uniform}
	Let $\delta \in (0,1]$, $0 < \la \leq \nor{C}$, $n, M \in \N$ and the \Nystrom{} centers be selected via uniform sampling. Let $\widehat{f}_{\la, M, \iter}$ be the FALKON estimator, after $\iter$ iterations (Def.~\ref{def:generalized-FALKON}) and let $\widetilde{f}_{\la, M}$ be the exact \Nystrom{} estimator in Eq.~\eqref{eq:base-nystrom}. When
	$$
	M \geq 5\left[1 + 14{\cal N}_\infty(\la)\right]\log \frac{8\kappa^2}{\la \delta},
	$$
	then, with probability $1 - 2\delta$,
	$$
	{\cal R}(\widehat{f}_{\la, M, \iter})^{1/2} ~~\leq~~ {\cal R}(\widetilde{f}_{\la, M})^{1/2} ~~ + ~~ 4\widehat{v} ~ e^{- \frac{\iter}{2}} ~ \sqrt{1 + \frac{9\kappa^2}{\la n} \log \frac{n}{\delta}},
	$$
	\et
	\bpr
	By applying Lemma~\ref{lm:ny-uni-controls-W} we have that
	$$\cond{W} \leq (e^{1/2}+1)^2/(e^{1/2}-1)^2,$$
	with probability $1-\delta$ under the condition on $M$. Then apply the computational oracle inequality in Thm.~\ref{thm:oracle-inequality-base} and take the union bound of the two events.
	\epr

	{\bf Theorem~\ref{thm:M-simple-rates}.}~{\em
		Under the same conditions of Thm.~\ref{thm:oracle-inequality-base}, the exponent $\nu$ in Thm.~\ref{thm:oracle-inequality-base} satisfies $\nu \geq 1/2$, with probability $1 - 2\delta$, when the \Nystrom{} centers are selected via uniform sampling (see Sect.~\ref{sect:gen-algo}), and
		$$
		M \geq 5\left[1 + \frac{14\kappa^2}{\la}\right]\log \frac{8\kappa^2}{\la \delta}.
		$$
	}
	\bpr
	It is a direct application of Thm.~\ref{thm:oracle-uniform}. Indeed note that ${\cal N}_\infty(\la) \leq \frac{\kappa^2}{\la}$ by definition.
	\epr
	
	\subsection{Main Result (II): computational oracle inequality for FALKON with leverage scores}
	
	\blm\label{lm:ny-als-controls-W}
	Let $\delta \in (0,1]$ and the matrix $W$ be as in Eq.~\ref{eq:gen-FALKON-problem} with $B$ satisfying Def.~\ref{def:matr-algo} and the \Nystrom{} centers selected via  $(q,\la_0,\delta)$-approximated leverage scores sampling (see Def.~\ref{def:emp-q-als} and discussion below), with $\la_0 = \frac{19\kappa^2}{n} \log \frac{n}{2\delta}$.
	When $\la_0 \leq \la \leq \nor{C}$, $n \geq 405 \kappa^2 \vee 67 \kappa^2 \log \frac{12\kappa^2}{\delta}$ and
	\eqal{\label{eq:ny-als-cond-M-final}
		M \geq 5\left[1 + 43q^2{\cal N}(\la)\right]\log \frac{8\kappa^2}{\la \delta},
	}
	then the following holds with probability $1-\delta$
	$$
	\cond{W} \leq \left(\frac{e^{1/2} + 1}{e^{1/2} - 1}\right)^2.
	$$
	\elm
	\bpr
	By Lemma~\ref{lm:characterization-W} we have that
	$$\cond{W} \leq \frac{1 + \nor{E}}{1 - \nor{E}},$$
	with the operator $E$ defined in the same lemma. By Lemma~\ref{lm:control-E} we have
	$$\nor{E} \leq \nor{\Gml^{-1/2}(\Cn - \Gm)\Gml^{-1/2}}.$$
	Lemma~\ref{lm:ny-als} proves that when the \Nystrom{} centers are selected via $q$-approximate leverage scores and $M$ satisfies Eq.~\eqref{eq:lev-nyst-beta} for a given parameter $\eta \in (0,1]$, then $\nor{\Gml^{-1/2}(\Cn - \Gm)\Gml^{-1/2}} \leq \eta$, with probability $1 - \delta$. In particular we select $\eta = \frac{2e^{1/2}}{e + 1}$. The condition on $M$ in Eq.~\eqref{eq:ny-als-cond-M-final} is derived by Eq.~\eqref{eq:lev-nyst-beta} by substituting $\eta$ with $\frac{2e^{1/2}}{e + 1}$.
	\epr

	\bt\label{thm:oracle-als}
	Let $\delta \in (0, 1], M, n \in \N$ and the \Nystrom{} centers be selected via $(q, \la_0,\delta)$-approximated leverage scores sampling (see Def.~\ref{def:emp-q-als} and discussion below), with $\la_0 = \frac{19\kappa^2}{n} \log \frac{n}{2\delta}$. Let $t \in \N$. Let $\widehat{f}_{\la, M, \iter}$ be the FALKON estimator, after $\iter$ iterations (Def.~\ref{def:generalized-FALKON}) and let $\widetilde{f}_{\la, M}$ be the exact \Nystrom{} estimator in Eq.~\eqref{eq:base-nystrom}. When $\la_0 \leq \la \leq \nor{C}$, $n \geq 405 \kappa^2 \vee 67 \kappa^2 \log \frac{12\kappa^2}{\delta}$ and
	$$
	M \geq 5\left[1 + 43q^2{\cal N}(\la)\right]\log \frac{8\kappa^2}{\la \delta},
	$$
	then, with probability $1 - 2\delta$,
	$$
	{\cal R}(\widehat{f}_{\la, M, \iter})^{1/2} ~~\leq~~ {\cal R}(\widetilde{f}_{\la, M})^{1/2} ~~ + ~~ 4\widehat{v} ~ e^{- \frac{\iter}{2}} ~ \sqrt{1 + \frac{9\kappa^2}{\la n} \log \frac{n}{\delta}},
	$$
	\et
	\bpr
	By applying Lemma~\ref{lm:ny-als-controls-W} we have that
	$$\cond{W} \leq (e^{1/2}+1)^2/(e^{1/2}-1)^2,$$
	with probability $1-\delta$ under the conditions on $\la, n, M$. Then apply the computational oracle inequality in Thm.~\ref{thm:oracle-inequality-base} and take the union bound of the two events.
	\epr
	
	{\bf Theorem \ref{thm:M-fast-rates}.}~{\em
		Under the same conditions of Thm.~\ref{thm:oracle-inequality-base}, the exponent $\nu$ in Thm.~\ref{thm:oracle-inequality-base} satisfies $\nu \geq 1/2$, with probability $1- 2\delta$, when\\
		\begin{enumerate}
			\item either \Nystrom{} uniform sampling (see Sect.~\ref{sect:gen-algo}) is used with $M \geq 70\left[1 + {\cal N}_\infty(\la)\right]\log \frac{8\kappa^2}{\la \delta}.$
			\item or \Nystrom{} $(q,\la_0, \delta)$-appr. lev. scores (see Sect.~\ref{sect:gen-algo}) is used, with $\la \geq \frac{19\kappa^2}{n} \log \frac{n}{2\delta}$, $n \geq 405 \kappa^2 \log \frac{12\kappa^2}{\delta},$~and
			$$
			M \geq 215\left[2 + q^2{\cal N}(\la)\right]\log \frac{8\kappa^2}{\la \delta}.
			$$
		\end{enumerate}  
	}
	\bpr
	It is a merge of Thm.~\ref{thm:oracle-uniform} and Thm.~\ref{thm:oracle-als}.
	\epr

	\subsection{Main Results (III): Optimal Generalization Bounds}
	
	First we recall the standard assumptions to study generalization rates for the non-parametric supervised learning setting, with square loss function. Then we provide Thm.~\ref{thm:general}, from which we obtain Thm.~\ref{thm:simple-rates} and Thm.~\ref{thm:fast-rates}.
	
	There exists $\kappa \geq 1$ such that $K(x,x) \leq \kappa^2$ for any $x \in \X$.  There exists $\fh \in \hh$, such that ${\cal E}(\fh) = \inf_{f \in \hh} {\cal E}(f)$. Moreover, we assume that $\rho(y|x)$ has sub-exponential tails, i.e. in terms of moments of $y$: there exist $\sigma, b$ satisfying $0 \leq \sigma \leq b$, such that, for any $x \in X$, the following holds
	\eqal{\label{eq:ass-noise}
		\mathbb{E} \left[\left|y - \fh(x)\right|^p ~ | ~ x \right] \leq \frac{1}{2} p! \sigma^2 b^{p-2}, \quad \forall p \geq 2.
	}
	Note that the assumption above is satisfied, when $y$ is supported in an interval or when it has sub-gaussian or sub-exponential tails.
	The last assumption is known as {\em source condition} \cite{caponnetto2007optimal}. There exist $r \in [1/2, 1]$ and $g \in \hh$, such that
	\eqal{\label{eq:ass-source}
		\fh = C^{r-1/2} g,
	}
	where $C$ is the correlation operator defined in Def.~\ref{def:ideal-op}. Finally define $R \geq 1$ such that $R \geq \nor{g}_\hh$. Note that assuming the existence of $\fh$, the source condition is always satisfied with $r = 1/2, g = \fh$ and $R = \max(1, \nor{\fh}_\hh)$, however if it is satisfied with larger $r$ it leads to faster learning rates.
	
	\bt\label{thm:general}
	Let $\delta \in (0,1]$. Let $n, \la, M$ satisfy
	$n \geq 1655\kappa^2 + 223 \kappa^2 \log \frac{24\kappa^2}{\delta}$,  $M \geq 334 \log \frac{192 n}{\delta}$ and $\frac{19\kappa^2}{n} \log \frac{24 n}{\delta} \leq \la \leq \nor{C}$.
	Let $\widehat{f}_{\la, M, \iter}$ be the FALKON estimator in Def.~\ref{def:generalized-FALKON}, after $\iter \in \N$ iterations.
	Under the assumptions in Eq.~\eqref{eq:ass-noise},~\eqref{eq:ass-source}, the following holds with probability at least $1 - \delta$,
	\eqal{\label{eq:final-bound-general}
		{\cal R}(\widehat{f}_{\la, M, \iter})^{1/2} \leq 6R \left(\frac{b\sqrt{{\cal N}_\infty(\la)}}{n} + \sqrt{\frac{\sigma^2{\cal N}(\la)}{n}}\right)\log\frac{24}{\delta} + 7 R \la^r,
	}
	\begin{enumerate}
		\item either, when the \Nystrom{} points are selected uniformly sampled (see Sect.~\ref{sect:gen-algo}) and
		\eqal{\label{eq:cond-ny-uni}
			M \geq 70\Big[1 ~+~ {\cal N}_\infty(\la)\Big] \log \frac{48\kappa^2}{\la\delta}, \qquad \iter \geq 2\log \frac{8(b + \kappa \nor{\fh}_\hh)}{R \la^r},
		}
		\item or, when the \Nystrom{} points are selected by means of $(q, \la_0, \delta)$-approximate leverage scores (see Sect.~\ref{sect:gen-algo}), with $q \geq 1$, $\la_0 = \frac{19\kappa^2}{n} \log \frac{48n}{\delta}$ and
		\eqal{\label{eq:cond-ny-als}
			M \geq 215\Big[1 ~+~ q^2 {\cal N}(\la)\Big]\log \frac{192 \kappa^2 n}{\la\delta}, \qquad \iter \geq 2\log \frac{8(b + \kappa \nor{\fh}_\hh)}{R \la^r}.
		}
	\end{enumerate}
	\et
	\bpr
	Let $\mu = \delta/4$.
	By Proposition~2 of \cite{rudi2015less}, under the assumptions in Eq.~\ref{eq:ass-noise} and Eq.~\ref{eq:ass-source},
	when $n \geq 1655\kappa^2 + 223 \kappa^2 \log \frac{6\kappa^2}{\mu}$, $~~ M \geq 334 \log \frac{48 n}{\mu}$, and $\frac{19\kappa^2}{n} \log \frac{6 n}{\mu} \leq \la \leq \nor{C},$ we have with probability $1-\mu$
	\eqals{
		{\cal R}(\widetilde{f}_{\la, M})^{1/2} \leq 6R \left(\frac{b\sqrt{{\cal N}_\infty(\la)}}{n} + \sqrt{\frac{\sigma^2{\cal N}(\la)}{n}}\right)\log\frac{6}{\mu} + 3R {\cal C}(M)^r + 3R \la^r,
	}
	where
	$$ {\cal C}(M) = \min \left\{ t > 0 ~~\middle|~~ (67 + 5 {\cal N}_{\infty}(t))\log \frac{12\kappa^2}{t \mu}  \leq M\right\},$$
	when the \Nystrom{} centers are selected with uniform sampling, otherwise
	$$ {\cal C}(M) = \min \left\{ \la_0 \leq  t \leq \nor{C} ~~\middle|~~  78 q^2 {\cal N}(t) \log \frac{48 n}{\mu} \leq M \right\},$$
	when the \Nystrom{} centers are selected via approximate sampling, with $\la_0 = \frac{19\kappa^2}{n} \log \frac{12 n}{\mu}$.
	In particular, note that ${\cal C}(M) \leq \la$, in both cases, when $M$ satisfies Eq.~\eqref{eq:cond-ny-uni} for uniform sampling, or Eq.~\eqref{eq:cond-ny-als} for approximate leverage scores.
	Now, by applying the computational oracle inequality in Thm.~\ref{thm:oracle-uniform}, for uniform sampling, or Thm.~\ref{thm:oracle-als}, for approximate leverage scores, the following holds with probability $1-2\mu$
	$$ R(\widehat{f}_{\la, M, \iter})^{1/2} \leq {\cal R}(\widetilde{f}_{\la, M})^{1/2} + 4\widehat{v} ~ e^{- \frac{\iter}{2}} ~ \sqrt{1 + \frac{9\kappa^2}{\la n} \log \frac{n}{\mu}},$$
	with $\widehat{v}^2 : = \frac{1}{n} \sum_{i=1}^n y_i^2$. In particular, note that, since we require $\la \geq \frac{19\kappa^2}{n} \log \frac{12 n}{\mu}$, we have
	$$4\sqrt{1 + \frac{9\kappa^2}{\la n} \log \frac{n}{\mu}} \leq 5.$$
	Now, we choose $\iter$ such that $5 \widehat{v} e^{-\iter/2} \leq R \la^r$, that is $t \geq 2 \log \frac{5 \widehat{v}}{R\la^r}$.
	The last step consists in bounding $\widehat{v}$ in probability. Since it depends on the random variables $y_1, \dots, y_n$ we bound it in the following way. By recalling that $|\fh(x)| = |\scal{K_x}{\fh}_\hh| \leq \nor{K_x}_\hh \nor{\fh}_\hh \leq \kappa \nor{\fh}_\hh$ for any $x \in \X$, we have
	$$\widehat{v} = \frac{1}{\sqrt{n}}\nor{\yn} \leq \sqrt{\sum_{i=1}^n \frac{(y_i - \fh(x_i))^2}{n}}  + \sqrt{\sum_{i=1}^n \frac{\fh(x_i)^2}{n}} \leq \sqrt{\sum_{i=1}^n \frac{(y_i - \fh(x_i))^2}{n}}  + \kappa\nor{\fh}_\hh.$$
	Since the training set examples $(x_i, y_i)_{i=1}^n$ are i.i.d. with probability $\rho$ we can apply the Bernstein inequality \cite{boucheron2004concentration} to the random variables $z_i = (y_i - \fh(x_i))^2 - s$, with $s = \mathbb{E} (y_i - \fh(x_i))^2$ (since $x_i, y_i$ are i.i.d. each $z_i$ has the same distribution and so the same expected value $s$).
	In particular, we need to bound the moments of $z_i$'s. By the assumption in Eq.~\ref{eq:ass-noise}, $z_i$ are zero mean and
	$$\mathbb{E} |z_i|^{2p} \leq \frac{1}{2}(2p)! \sigma^2b^{2p-2} \leq \frac{1}{2}p! (4\sigma b)^2(4b^2)^{p-2}, \qquad p \geq 2$$
	and so, by applying the Bernstein inequality, the following holds with probability $1-\mu$
	$$\left|\sum_{i=1}^n \frac{z_i}{n}\right| \leq \frac{8b^2\log\frac{2}{\mu}}{3n} + \sqrt{\frac{8\sigma^2 b^2\log\frac{2}{\mu}}{n}} \leq \frac{1}{4} b^2,$$
	where the last step is due to the fact that we require $n \geq 223 \kappa^2 \log \frac{6}{\mu}$, that $b \geq \sigma$ and that $\kappa \geq 1$ by definition.
	So, by noting that $s \leq \sigma^2 \leq b^2$ (see Eq.~\ref{eq:ass-noise}), we have
	$$\widehat{v} ~\leq~ \kappa \nor{\fh}_\hh + \sqrt{s + \sum_{i=1}^n \frac{z_i}{n}} ~\leq~ \kappa \nor{\fh}_\hh + \sqrt{s} + \frac{1}{2}b ~\leq~ \frac{3}{2} b + \kappa \nor{\fh}_\hh,$$
	with probability at least $1 - \mu$.
	Now by taking the intersection of the three events, the following holds with probability at least $1- 4\mu$
	$${\cal R}(\widehat{f}_{\la, M, \iter})^{1/2} \leq 6R \left(\frac{b\sqrt{{\cal N}_\infty(\la)}}{n} + \sqrt{\frac{\sigma^2{\cal N}(\la)}{n}}\right)\log\frac{6}{\mu} + 7R \la^r.
	$$
	\epr
	
	Now we provide the generalization error bounds for the setting where we only assume the existence of $\fh$.
	
	{\bf Theorem.~\ref{thm:simple-rates}}~~{\em
		Let $\delta \in (0,1]$. Let the outputs $y$ be bounded in $[-\frac{a}{2}, \frac{a}{2}]$, almost surely, with $a > 0$. For any $n \geq \max(\frac{1}{\nor{C}}, 82\kappa^2 \log \frac{373\kappa^2}{\sqrt{\delta}})^2$ the following holds. When
		$$\la = \frac{1}{\sqrt{n}}, \qquad M \geq 5(67 + 20 \sqrt{n})\log\frac{48\kappa^2 n}{ \delta}, \qquad \iter ~\geq~ \frac{1}{2}\log(n) ~+~ 5 + 2\log (a + 3\kappa),$$
		then with probability $1-\delta$,
		$$ {\cal R}(\widehat{f}_{\la, M, t}~) \leq \frac{c_0 \log^2 \frac{24}{\delta}}{\sqrt{n}},$$
		where $\widehat{f}_{\la, M, t}$ is the FALKON estimator in Def.~\ref{def:generalized-FALKON} (see also Sect.~\ref{sect:FALKON} Alg.~\ref{algo:main}) with \Nystrom{} uniform sampling, and the constant $c_0 = 49\nor{\fh}_\hh^2(1 + a\kappa + 2\kappa^2\nor{\fh}_\hh)^2.$
	}
	\bpr
	Here we assume $y \in [-\frac{a}{2}, \frac{a}{2}]$ a.s., so Eq.~\ref{eq:ass-noise} is satisfied with $\sigma = b = a + 2 \kappa \nor{\fh}_\hh$, indeed
	$$ \mathbb{E}[|y - \fh(x)|^p~|~ x] ~\leq~ \mathbb{E}[2^{p-1}|y|^p ~|~ x] + 2^{p-1}|\fh(x)|^p ~\leq~ \frac{1}{2} (a^p + 2^p \kappa^p\nor{\fh}_\hh^p) \leq \frac{1}{2} p! (a + 2 \kappa\nor{\fh}_\hh)^p,$$
	where we used the fact that $|\fh(x)| =|\scal{K_x}{\fh}_\hh| \leq \nor{K_x}\nor{\fh} \leq \kappa \nor{\fh}_\hh$.
	Moreover, Eq.~\eqref{eq:ass-source} is satisfied with $r = 1/2$ and $g = \fh$, while $R = \max(1, \nor{\fh}_\hh)$.
	
	To complete the proof we show that the assumptions on $\la, M, n$ satisfy the condition required by Thm.~\ref{thm:general}, then we apply it and derive the final bound.
	Set $\la = n^{-1/2}$ and define $n_0 = \max(\nor{C}^{-1}, 82\kappa^2 \log \frac{373\kappa^2}{\sqrt{\delta}})^2$. The condition $n \geq n_0$, satisfies the condition on $n$ required by Thm.~\ref{thm:general}. Moreover both $\la = n^{-1/2}$ and $M \geq 75 ~\sqrt{n}~\log\frac{48\kappa^2 n}{ \delta}$ satisfy respectively the conditions on $\la, M$ required by Thm.~\ref{thm:general}, when $n \geq n_0$. Finally note that the condition on $\iter$ implies the condition required by Thm.~\ref{thm:general}, indeed, since $R = \max(1, \nor{f}_\hh)$, we have $a/R \leq a$ and $\nor{\fh}_\hh/R \leq 1$, so
	\eqals{
		2\log \frac{8(a + \kappa \nor{\fh}_\hh)}{R \la^r} & ~=~ \log \left[64\left(\frac{a}{R} + \frac{3\kappa\nor{\fh}_\hh}{R}\right)^2 \sqrt{n}\right] \\
		& ~\leq~ \log (64(a + 3\kappa)^2 \sqrt{n}) ~\leq~ \log 64 + \frac{1}{2}\log n + 2 \log(a + 3\kappa).
	}
	So, by applying Thm.~\ref{thm:general} with $R, r$ defined as above and recalling that ${\cal N}(\la) \leq {\cal N}_\infty(\la) \leq \frac{\kappa^2}{\la}$, we have
	\eqals{
		{\cal R}(\widehat{f}_{\la, M, \iter})^{1/2} &\leq 6R \left(\frac{b\sqrt{{\cal N}_\infty(\la)}}{n} + \sqrt{\frac{\sigma^2{\cal N}(\la)}{n}}\right)\log\frac{24}{\delta} + 7 R \la^r \leq 6R \left(\frac{b\kappa}{\sqrt{\la} n} + \frac{\sigma \kappa}{\sqrt{\la n}}\right)\log\frac{24}{\delta} + 7 R \la^{\frac12} \\
		&= 6R b \kappa (1 + n^{-1/2}) n^{-1/4} \log\frac{24}{\delta} + 7 R n^{-1/4}  \leq \frac{7R(b \kappa + 1) \log \frac{24}{\delta}}{n^{1/4}}.
	}
	with probability $1-\delta$. For the last step we used the fact that $b = \sigma$, that $6(1 + n^{-1/2}) \leq 7$, since $n \geq n_0$, and that $\log \frac{24}{\delta} > 1$.
	\epr
	
	To state the result for fast rates, we need to define explicitly the {\em capacity condition} on the intrinsic dimension. There exists $Q > 0, \gamma \in (0,1]$ such that
	\eqal{\label{eq:ass-capacity}
		{\cal N}(\la) \leq Q^2 \la^{-\gamma}, \quad \forall \la \geq 0.
	}
	Note that, by definition of ${\cal N}(\la)$, the assumption above is always satisfied with $Q = \kappa$ and $\gamma = 1$.
	
	{\bf Theorem \ref{thm:fast-rates}}~~{\em
		Let $\delta \in (0,1]$. Let the outputs $y$ be bounded in $[-\frac{a}{2}, \frac{a}{2}]$, almost surely, with $a > 0$. Under the assumptions in Eq.~\ref{eq:ass-source},~\ref{eq:ass-capacity} and $n \geq \nor{C}^{-s} \vee ~\left(\frac{102 \kappa^2s}{s-1} \log \frac{912}{\delta}\right)^{\frac{s}{s-1}}$, with $s = 2r + \gamma$, the following holds. When
		$$\la = n^{-\frac{1}{2r + \gamma}}, \qquad \iter ~\geq~ \log(n) ~+~ 5 + 2\log(a + 3\kappa^2),$$
		\begin{enumerate}
			\item and either \Nystrom{} uniform sampling is used with
			\eqal{\label{eq:fast-rates-M-uni}
				M \geq 70 \left[1 ~+~ {\cal N}_\infty(\la)\right]~\log \frac{8\kappa^2}{\la \delta},
			}
			\item and or \Nystrom{} $(q, \la_0, \delta)$-approximate leverage scores (Def.~\ref{def:emp-q-als}), with $q \geq 1$, $\la_0 = \frac{19\kappa^2}{n} \log \frac{48n}{\delta}$ and
			\eqal{\label{eq:fast-rates-M-als}
				M \geq 215\left[1 ~+~ q^2{\cal N}(\la)\right]~\log \frac{8\kappa^2}{\la \delta},
			}
		\end{enumerate}
		then with probability $1-\delta$,
		$$ {\cal R}(\widehat{f}_{\la, M, t}) ~\leq~ c_0 \log^2 \frac{24}{\delta}~~n^{-\frac{2r}{2r+\gamma}},$$
		where $\widehat{f}_{\la, M, t}$ is the FALKON estimator in Sect.~\ref{sect:FALKON} (Alg.~\ref{algo:main}). In particular $n_0, c_0$ do not depend on $\la, M, n$ and $c_0$ do not depend on $\delta$.
	}
	\bpr
	The proof is similar to the one for the slow learning rate (Thm.~\ref{thm:simple-rates}), here we take into account the additional assumption in Eq.~\eqref{eq:ass-source},~\ref{eq:ass-capacity} and the fact that $r$ may be bigger than $1/2$. Moreover we assume $y \in [-\frac{a}{2}, \frac{a}{2}]$ a.s., so Eq.~\ref{eq:ass-noise} is satisfied with $\sigma = b = a + 2 \kappa \nor{\fh}_\hh$, indeed
	$$ \mathbb{E}[|y - \fh(x)|^p~|~ x] ~\leq~ \mathbb{E}[2^{p-1}|y|^p ~|~ x] + 2^{p-1}|\fh(x)|^p ~\leq~ \frac{1}{2} (a^p + 2^p \kappa^p\nor{\fh}_\hh^p) \leq \frac{1}{2} p! (a + 2 \kappa\nor{\fh}_\hh)^p,$$
	where we used the fact that $|\fh(x)| =|\scal{K_x}{\fh}_\hh| \leq \nor{K_x}\nor{\fh} \leq \kappa \nor{\fh}_\hh$.
	
	To complete the proof we show that the assumptions on $\la, M, n$ satisfy the required conditions to apply Thm.~\ref{thm:general}. Then we apply it and derive the final bound.
	Set $\la = n^{-1/(2r + \gamma)}$ and define $n_0 = \nor{C}^{-s} \vee ~\left(\frac{102 \kappa^2s}{s-1} \log \frac{912}{\delta}\right)^{\frac{s}{s-1}}$, with $s = 2r + \gamma$. Since $1 < s \leq 3$, the condition $n \geq n_0$, satisfies the condition on $n$ required to apply Thm.~\ref{thm:general}. Moreover, for any $n \geq n_0$, both $\la = n^{-1/(2r+\gamma)}$ and $M$ satisfying Eq.~\eqref{eq:fast-rates-M-uni} for \Nystrom{} uniform sampling, and Eq.~\ref{eq:fast-rates-M-als} for \Nystrom{} leverage scores, satisfy respectively the conditions on $\la, M$ required to apply Thm.~\ref{thm:general}. Finally note that the condition on $\iter$ implies the condition required by Thm.~\ref{thm:general}, indeed, since $2r/(2r+\gamma)\leq 1$,
	\eqals{
		2\log \frac{8(b + \kappa \nor{\fh}_\hh)}{R \la^r} & ~=~ \log \left[64\left(\frac{a}{R} + \frac{3\kappa\nor{\fh}_\hh}{R}\right)^2 n^{\frac{2r}{2r+\gamma}}\right] \\
		& ~\leq~ \log 64  + 2\log \frac{a + 3\kappa\nor{\fh}_\hh}{R} +  \frac{2r}{2r+\gamma} \log n \\
		& ~\leq~ \log 64  + 2\log \frac{a + 3\kappa\nor{\fh}_\hh}{R} +  \log n,\\
		& ~\leq~ \log 64  + 2\log(a + 3\kappa^2) +  \log n.
	}
	where the last step is due to the fact that $a/R \leq 1$ and $\nor{f_h}_\hh/R \leq \nor{C^{r-1/2}} \leq \nor{C}^{1/2} \leq \kappa$, since $R := \max(1,\nor{g}_\hh)$, and $\nor{\fh}_\hh \leq \nor{C^{r-1/2}} \nor{g}_\hh$, by definition.
	So, by applying Thm.~\ref{thm:general} with $R, r$ defined as above and recalling that ${\cal N}_\infty(\la) \leq \frac{\kappa^2}{\la}$ by construction and that ${\cal N}(\la) \leq Q^2 \la^{-\gamma}$ by the capacity condition in Eq.~\eqref{eq:ass-capacity}, we have
	\eqals{
		{\cal R}(\widehat{f}_{\la, M, \iter})^{1/2} &\leq 6R \left(\frac{b\sqrt{{\cal N}_\infty(\la)}}{n} + \sqrt{\frac{\sigma^2{\cal N}(\la)}{n}}\right)\log\frac{24}{\delta} + 7 R \la^r \leq 6R \left(\frac{b\kappa}{\sqrt{\la} n} + \frac{Q\sigma}{\sqrt{\la^{\gamma} n}}\right)\log\frac{24}{\delta} + 7 R \la^{r} \\
		&= 6Rb\left(\kappa n^{-\frac{r + \gamma - 1/2}{2r+\gamma}} + Q\right) n^{-\frac{r}{2r+\gamma}} \log\frac{24}{\delta} + 7 R n^{-\frac{r}{2r+\gamma}}  \\
		& \leq ~ 7R(b(\kappa + Q) + 1) \log \frac{24}{\delta}~~n^{-\frac{r}{2r+\gamma}}.
	}
	with probability $1-\delta$. For the last step we used the fact that $b = \sigma$, that $r + \gamma - 1/2 \geq 0$, since $r \geq 1/2$ by definition, and that $\log \frac{24}{\delta} > 1$.
	\epr
	
	\section{Longer comparison with previous works} \label{sect:long-comparison}
	
	In the literature of KRR there are some papers that propose to solve Eq.~\ref{eq:krls} with iterative preconditioned methods \cite{fasshauer2012stable,avron2016faster, cutajar2016preconditioning, gonen2016solving, ma2017diving}.
	In particular the one of \cite{fasshauer2012stable} is based, essentially, on an incomplete singular value decomposition of the kernel matrix. Similarly, the ones proposed by \cite{gonen2016solving, ma2017diving} are based on singular value decomposition obtained via randomized linear algebra approaches. The first covers the linear case, while the second deals with the kernel case. \cite{avron2016faster, cutajar2016preconditioning} use a preconditioner based on the solution of a randomized projection problem based respectively on random features and \Nystrom{}.
	
	While such preconditioners are suitable in the case of KRR, their computational cost becomes too expensive when applied to the random projection case. Indeed, performing an incomplete svd of the matrix $\Knm$ even via randomized linear algebra approaches would require $O(nMk)$ where $k$ is the number of singular values to compute. To achieve a good preconditioning level (and so having $t \approx \log n$) we should choose $k$ such that $\sigma_k(\Knm) \approx \la$. When the kernel function is bounded, without further assumptions on the eigenvalue decay of the kernel matrix, we need $k \approx \la^{-1}$ \cite{caponnetto2007optimal,rudi2015less}. Since randomized projection requires $\la = n^{-1/2}$, $M =O(\sqrt{n})$ to achieve optimal generalization bounds, we have $k \approx \sqrt{n}$ and so the total cost of the incomplete svd preconditioner is $O(n^2)$. On the same lines, applying the preconditioner proposed by \cite{avron2016faster,cutajar2016preconditioning} requires $O(nM^2)$ to be computed and there is no natural way to find a similar sketched preconditioner as the one in Eq.~\eqref{eq:B-base} in the case of \cite{avron2016faster}, with reduced computational cost. In the case of \cite{cutajar2016preconditioning}, the preconditioner they use is exactly the matrix $H^{-1}$, whose computation amounts to solve the original problem in Eq.~\eqref{eq:base-nystrom} with direct methods and requires $O(nM^2)$.

	A similar reasoning hold for methods that solve the \Nystrom{} linear system \eqref{eq:base-nystrom} with iterative approaches \cite{dai2014scalable,camoriano2016nytro,tu2016large}. Indeed on the positive side, they have a computational cost of $O(nM t)$. However they are affected by the poor conditioning of the linear system in Eq.~\ref{eq:base-nystrom}. Indeed, even if $H$ or $\Km$ in Eq.~\ref{eq:base-nystrom} are invertible, their condition number can be arbitrarily large (while in the KRR case it is bounded by $\la^{-1}$), and so many iterations are often needed to achieve optimal generalization (E.g. by using early stopping in \cite{camoriano2016nytro} they need  $t \approx \la^{-1}$).

	
	
\section{MATLAB Code for FALKON}\label{sect:matlab}

\begin{algorithm}[t]
	 {\small	
	\caption{Complete {\tt MATLAB} code for FALKON. It requires $O(nMt + M^3)$ in time and $O(M^2)$ in  memory. See Sect.~\ref{sect:FALKON} for more details, and Sect.~\ref{sect:teo} for theoretical properties.
		\label{algo:main-MATLAB}}
	\vspace{0.1cm}
	\begin{flushleft}
		{\bf Input:} Dataset $X = (x_i)_{i=1}^n \in \R^{n \times D}, \hat y = (y_i)_{i=1}^n \in \R^n$, $M \in \N$ numbers of \Nystrom{} centers to select, $\textrm{lev\_scores} \in \R^n$ approximate leverage scores (set $\textrm{lev\_scores = [~]}$ for selecting \Nystrom{} centers via uniform sampling), function $\textrm{KernelMatrix}$ computing the Kernel matrix of two sets of points, regularization parameter $\la$, number of iterations $\iter$.\\
		{\bf Output:} \Nystrom{} coefficients $\alpha$.
	\end{flushleft}
}	
 {\tiny
	\begin{center}
		\begin{verbatim}
function alpha = FALKON(X, Y, lev_scores, M, KernelMatrix, lambda, t)
    n = size(X,1);
    [C, D] = selectNystromCenters(X, lev_scores, M, n);

    KMM = KernelMatrix(C,C);
    T = chol(D*KMM*D + eps*M*eye(M));
    A = chol(T*T'/M + lambda*eye(M));

        function w = KnMtimesVector(u, v)
            w = zeros(M,1); ms = ceil(linspace(0, n, ceil(n/M)+1));
            for i=1:ceil(n/M)
                Kr = KernelMatrix( X(ms(i)+1:ms(i+1),:), C );
                w = w + Kr'*(Kr*u + v(ms(i)+1:ms(i+1),:));
            end
        end

        function w = BHB(u)
            w = A'\(T'\(KnMtimesVector(T\(A\u), zeros(n,1))/n) + lambda*(A\u));
        end

    r = A'\(T'\KnMtimesVector(zeros(M,1), Y/n));

    beta = conjgrad(@BHB, r, t);
    alpha = T\(A\beta);
end

function beta = conjgrad(funA, r, tmax)
    p = r; rsold = r'*r; beta = zeros(size(r,1), 1);

    for i = 1:tmax
        Ap = funA(p);
        a = rsold/(p'*Ap);
        beta = beta + a*p;
        r = r - a*Ap; rsnew = r'*r;
        p = r + (rsnew/rsold)*p;
        rsold = rsnew;
    end
end

function [C, D] = selectNystromCenters(X, lev_scores, M, n)
    if isempty(lev_scores) %Uniform Nystrom
        D = eye(M);
        C = X(randperm(n,M),:);
    else % Appr. Lev. Scores Nystrom
        prob = lev_scores(:)./sum(lev_scores(:));
        [count, ind] = discrete_prob_sample(M, prob);
        D = diag(1./sqrt(n*prob(ind).*count));
        C = X(ind,:);
    end
end

function [count, ind] = discrete_prob_sample(M, prob)
    bins = histcounts(rand(M,1), [0; cumsum(prob(:))]);
    ind = find(bins > 0);
    count = bins(ind);
end
		
		\end{verbatim}
	\end{center}
}
\vspace{-0.35cm}

\end{algorithm}	
	
\end{document}